\documentclass[journal]{IEEEtran}
\usepackage{amsmath}
\usepackage{graphicx}
\usepackage[mathletters]{ucs}
\usepackage{caption}
\usepackage[utf8x]{inputenc}
\usepackage{subcaption}

\begin{document}

\title{The Computerized Classification of Micro-Motions in
the Hand using Waveforms from Mobile Phone Videos}

\author{Ranjani~Ramesh
\thanks{Ranjani Ramesh is a 12th grade student and National Public School, Indiranagar, Bangalore, Karnataka 560008, India. Corresponding Author email: ranjani2122appls@gmail.com}
\thanks{Manuscript submitted October 13, 2021.}}

\maketitle

\begin{abstract}
Our hands reveal important information such as the pulsing of our veins which help us
determine the blood pressure, tremors indicative of motor control, or neuro-degenerative
disorders such as Essential Tremor or Parkinson’s Disease. The Computerized
Classification of Micro-Motions in the Hand using Waveforms from Mobile Phone Videos
is a novel method that uses Eulerian Video Magnification, Skeletonization, Heatmapping,
and the kNN machine learning model to detect the micro-motions in the
human hand, synthesize their waveforms, and classify these.
The pre-processing is achieved by using Eulerian Video Magnification, Skeletonization,
and Heat-mapping to magnify the micro-motions, landmark essential features of the
hand, and determine the extent of motion, respectively. Following pre-processing, the
visible motions are manually labeled by appropriately grouping pixels to represent a
particular label correctly. These labeled motions of the pixels are converted into
waveforms. Finally, these waveforms are classified into four categories - hand/finger
movements, vein movement, background motion, and movement of the rest of the body
due to respiration using the kNN model. The final accuracy obtained was around
92.48\%.
\end{abstract}


\IEEEpeerreviewmaketitle

\section{Introduction}
\label{Sec1}
%
%
%
%
\IEEEPARstart{D}{igital} media, particularly videos, play an essential role in the documentation of
information, in today’s world. With the help of computer vision and artificial intelligence,
important data stored in videos can be extracted and used in multiple domains with the
help of Computer Vision and AI. The videos of the hands of human beings reveal
essential information in the form of micro - motions such as the throbbing of the veins
due to the pulse or tremor components. The remote monitoring of vitals such as pulse
and respiration rates of a person and remote detection of neuro-degenerative diseases
such as Parkinson’s disease are essential in times such as the present COVID - 19
pandemic.

In this spirit, this paper proposes a computerized method to detect and classify these
micro motions present in the hand from mobile phone videos using video magnification,
heat-mapping, skeletonization, and deep-learning. This method aims to provide an atlas
or a guide for the different kinds of motions occurring in the human hand and thus help
in the computerized detection of specific motions. It can be used as an initial step
towards remote detection of movement disorders.

Previously proposed methods, such as Williams et al.’s proposition, which details a
method to aid in clinical detection of Parkinson’s Disease (PD) using a computerized
magnification of tremors in atremulous PD patients~\cite{Williams}, and Hillegondsberg et al.’s
approach~\cite{Hillegondsberg}, which aims to enhance the fasciculations in Amyotrophic Lateral Sclerosis
(ALS) patients using Eulerian Video Magnification~\cite{Wu}–\cite{Davis} have 
shown promising results in improving the accuracy of detection of diseases by specialists. 
However, a basic framework that can be used for easy detection of small motions in the hand is lacking.

This paper details an automated computerized method wherein videos of hands taken
using standard smartphones are simultaneously subjected to skeletonization and
Eulerian Video Magnification. The magnified result is heat-mapped, and the obtained
output is overlapped with the skeletonized output from the first step. This step results in
an output consisting of key points from the skeletonization and movement coloring from
the heatmap. The intensity of the color of the heatmap and the key points are used to
locate micro - motions in the hand. Waveforms of the motion of these pixels of interest
are created. These waveforms are then divided into training and testing data (7:3) and
passed into a classifier, which results in the automatic classification of the motion of the
pixel.

 

\section{Related Work}
\label{Sec2}

\IEEEpubidadjcol
\subsection{Eulerian Video Magnification for Parkinson’s Disease tremor detection}
\label{Sec2_1}

The paper ‘Seeing the unseen: Could Eulerian video magnification aid clinician detection of sub-clinical Parkinson’s tremor?’, by Williams et al.~\cite{Williams} is the cornerstone and main inspiration for my research. This paper describes an experiment to detect Parkinson’s Disease (PD) tremors in atremulous hands of patients using Eulerian Video Magnification. In this paper, the authors first described their data acquisition process. This process of collecting data is crucial since such curated databases of videos of hands, especially PD patients’ hands, are rarely available. The data acquired
is then subjected to Eulerian Video Magnification. The magnification process aims to magnify all sub-clinical tremors in the hand. The tremors in the hand consist of a mechanical component and a central component. The mechanical component of the hand consists of the natural frequency and a mechanical - reflex component which is caused due to increased mono-synaptic reflex. The central component consists of atypical oscillatory activity which occurs along the motor system of the human body. Tremors in regular subjects consist of both the mechanical and the central components, while the tremors in PD patients consist primarily of the central component. This experiment noted that the Eulerian Magnification magnified both the central and mechanical tremor components. This process proved to be useful as it provided greater magnified tremor in the hands of PD patients due to extra central tremor components in their hands. Thus, clinicians classified a significant proportion of PD patients correctly from the Eulerian video-magnified result (p $<$ 0.003).

\subsection{Usage of “Lazy Learning” Algorithms for Classification of Oscillometric Waveforms}
\label{Sec2_2}
\begin{figure}[ht]
    \centering
    \includegraphics[width = \linewidth]{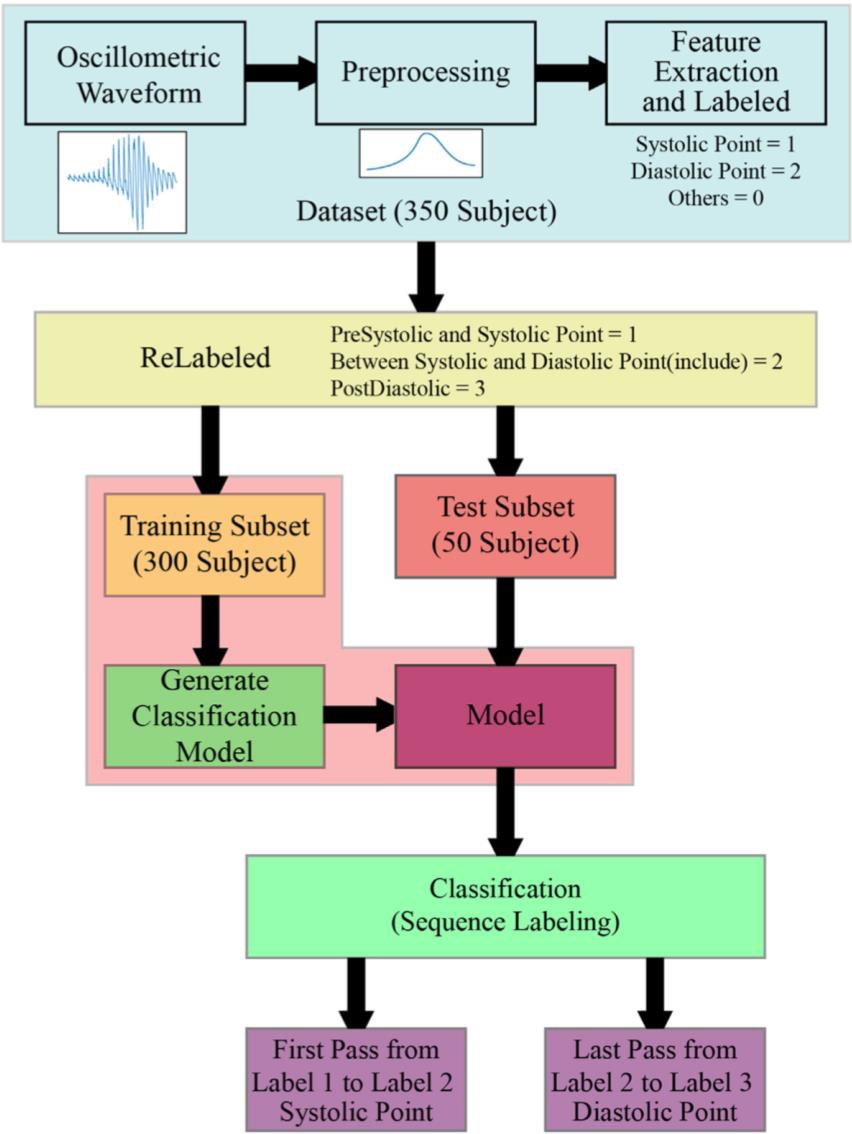}
    \caption{Flowchart Representation of Pipeline of Method detailed in “A novel blood pressure estimation method based on classification of oscillometric waveforms using machine-learning methods”.~\cite{Alghamdi}}
    \label{Fig1}
\end{figure}

Alghamdi et al.~\cite{Alghamdi} detail a method that predicts blood pressure using Oscillometric waveforms. The technique uses the BP Data-UNSW data set, which consists of 350 Non-Invasive Blood Pressure measurement records. The data was split into 300 records for training and 50 for testing. After appropriate pre-processing of the waveform images, features are extracted and are re-labeled as:
\begin{itemize}
\item 1 – beats before systolic
\item 2 – beats with systolic and diastolic points (including diastolic)
\item 3 – beats after diastolic point
\end{itemize}
as seen in Figure~\ref{Fig1}. The models used to predict the blood pressure were the k-Nearest Neighbors, Weighted k-Nearest Neighbors, and Bagged Trees. These models are referred to as “lazy learning” algorithms. An important inference derived from this paper is that lazy learning algorithms are as well-equipped as deep learning algorithms in such waveform analysis and classification tasks. Even though deep learning algorithms such as Deep Belief Network, Deep Neural Network and Long-Short Term Memory, and Recurrent Neural Networks (LSTM-RNN) have been preferred over models such as the kNN, Decision tree models, etc., this method has shown that models like the kNN model are well-equipped to tackle waveform classification tasks. This artcle also shows that the accuracy of kNN and wkNN are similar, and the accuracy of the Bagged Trees model is lower than these two models. This paper helped to determine an appropriate model for the task of classification of waveforms of different motions in the hand as detailed in Section~\ref{Sec3_3}.

\subsection{Lagrangian Motion Magnification}
\label{Sec2_3}
Liu et al.~\cite{Liu} detail a method of magnification of small motions. This paper proposes a method wherein the motions that must be amplified are selected by creating similar position and motion pixel clusters. In contrast to Eulerian Video Magnification, detailed in Section~\ref{Sec3_1}, this method follows the Lagrangian perspective wherein the trajectories of single particles are determined for accurate motion estimation, which is an essential step in magnifying the user-specified clusters. Subsequently, after the amplification of motion, 'holes' are formed. 

These holes are filled using texture synthesis methods. The main highlights of this method are:
\begin{itemize}
\item Its robust video registration helps to magnify only the tiny motions in the video and not the motions caused by the moving of the recording device
\item Its determination of feature trajectories by feature point detection
\item Its robust clustering of feature points trajectories
\end{itemize}

However, compared to Eulerian Video Magnification (Section~\ref{Sec3_1}), this method is computationally expensive due to its Lagrangian perspective. Additionally, it is more difficult to make complicated motions free of artifacts in this method~\cite{Rubinstein}. Due to these shortcomings of this method, Eulerian magnification was chosen instead.

\section{Background}
\label{Sec3}
\subsection{Eulerian Video Magnification}
\label{Sec3_1}
Eulerian Video Magnification~\cite{Wu},\cite{Davis},\cite{Burt} was chosen to magnify the tiny motions present in the hands of humans. Humans lack the ability to process small spatio-temporal changes through the naked eye. The minute motions of interest in the hand are below the naked eye’s sensitivity threshold, and thus, Eulerian Video Magnification is used to amplify these motions. This method was chosen as it has shown to produce good results in similar projects and works as detailed in Section 3.1. This process of magnification is extremely important in this pipeline as it magnifies all the micro - motions in the hand which is the basis to detect and classify these micro-motions.

This magnification method uses the Eulerian idea of fluid flow, which analyzes the changes of the properties of the fluid over time as opposed to the Lagrangian point of view, which analyzes the position and motion of individual particles, which is highly cumbersome.

Eulerian Video Magnification follows the pipeline of spatial processing by building a Laplacian pyramid using a Gaussian pyramid, and subsequently applying a band-passing filter to extract the frequency band of interest and magnification. The band-passing takes in all frequencies and passes those which are within a user - specified range while rejecting the frequencies outside the range, which is how the frequency band of interest is extracted.

\subsubsection{Spatial Processing}
\label{Sec3_1_1}
\begin{figure}[ht]
    \centering
    \includegraphics[width = \linewidth]{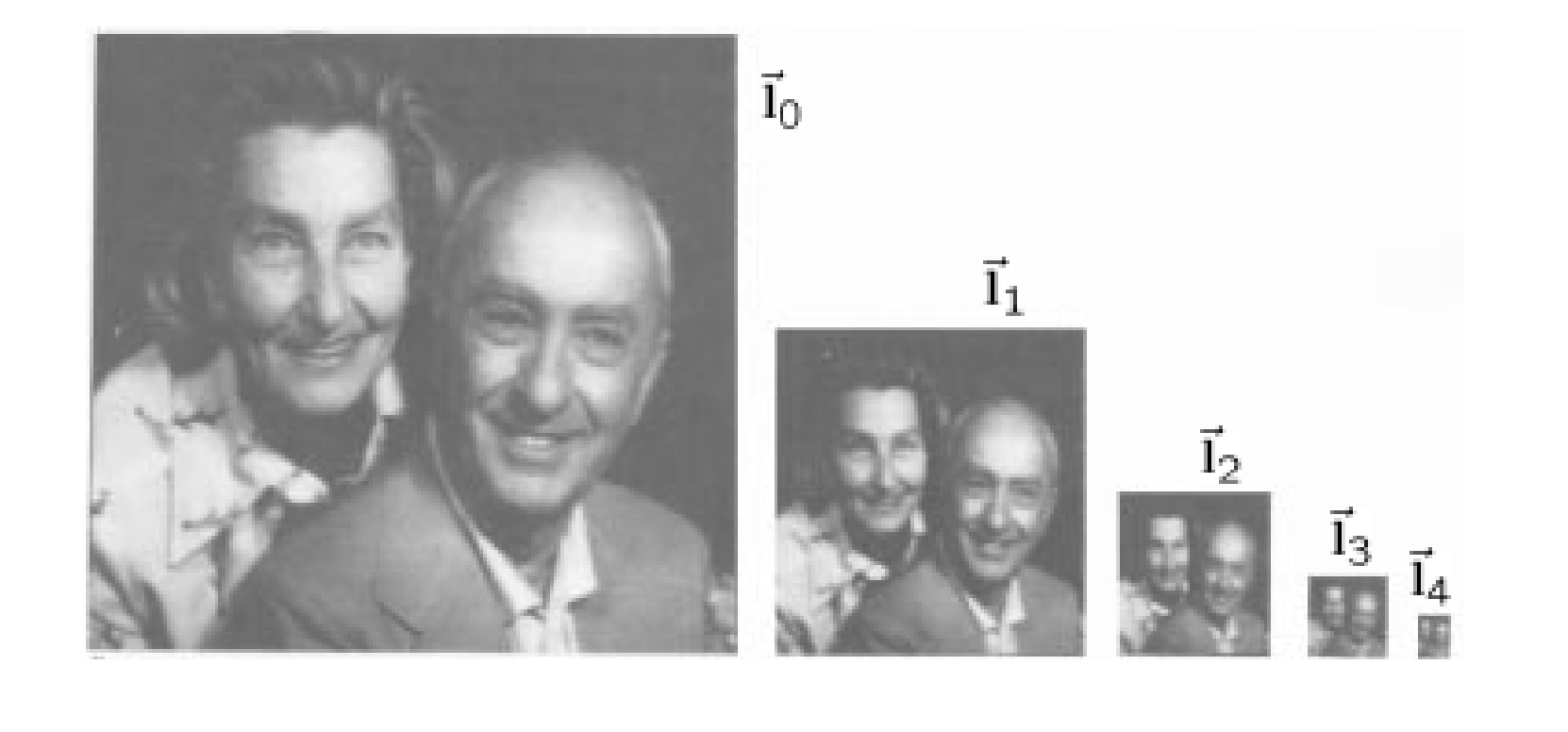}
    \caption{Gaussian Pyramid \\"http://www.cs.toronto.edu/$\sim$jepson/csc320/notes/\\pyramids.pdf"}
    \label{Fig2}
\end{figure}

\begin{figure}[ht]
    \centering
    \includegraphics[width = \linewidth]{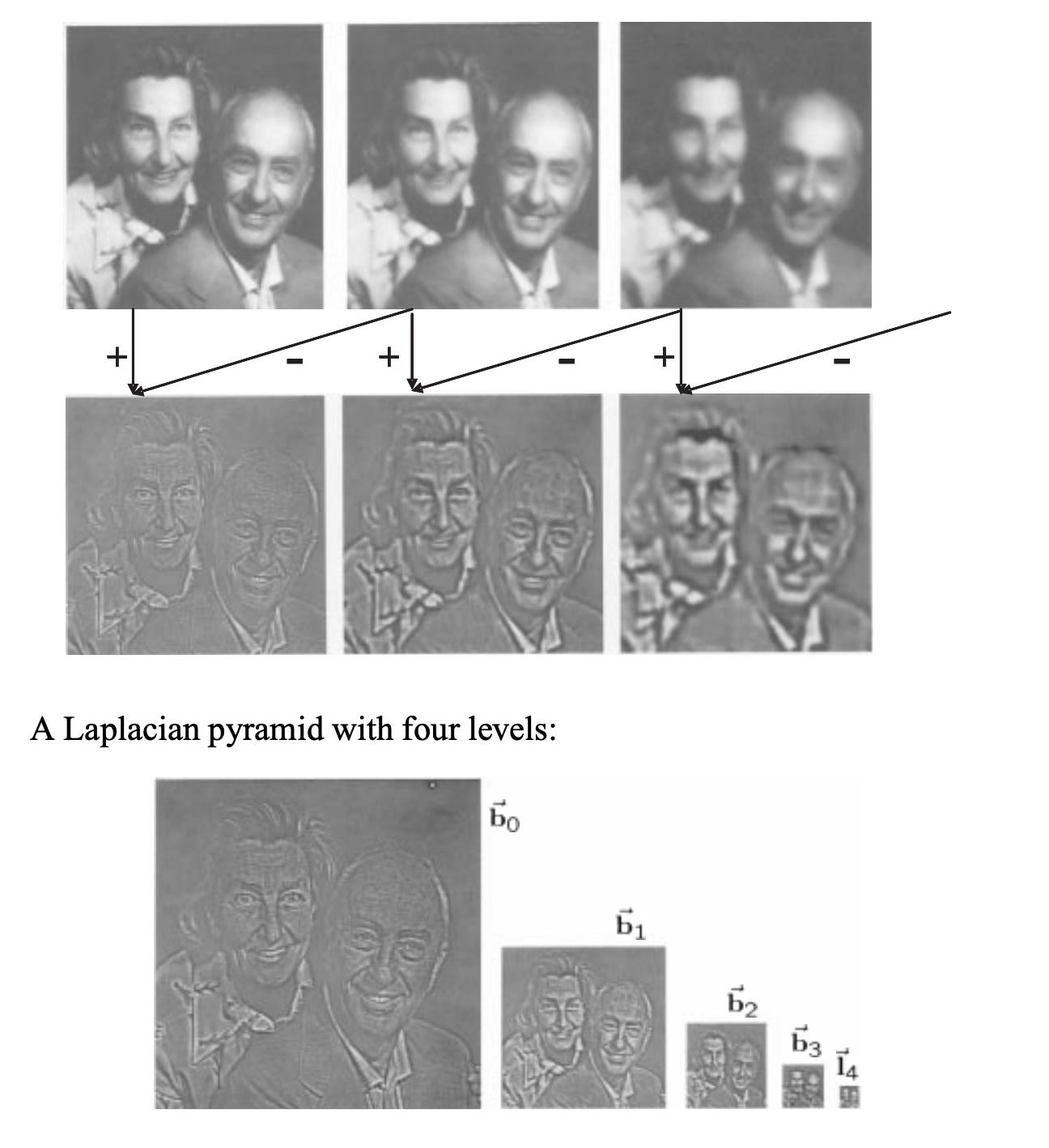}
    \caption{Laplacian Pyramid \\"http://www.cs.toronto.edu/$\sim$jepson/csc320/notes/\\pyramids.pdf"}
    \label{Fig3}
\end{figure}

Spatial processing [9] is accomplished by using a Laplacian pyramid in order to extract minute movements between successive frames. The Laplacian pyramid is created by first building a Gaussian Pyramid and then using its levels to build the Laplacian pyramid eventually. The pyramids help in the extraction of tiny motions since they minimize and maximize the images to different scales. The large-scale images help detect smaller motions and features while the small-scale images detect larger motions and features. This is depicted in Figs~\ref{Fig2},~\ref{Fig3}.

This section details the process of building both the Gaussian and the Laplacian pyramids.

\paragraph{Gaussian Pyramid}
Consider an image represented by array $I_0$. $I_0$ is the bottom level of the Gaussian Pyramid to be built. It undergoes low pass filtering and results in the obtainment of $I_1$, which is smaller and lower in resolution than $I_0$. A low-pass filter is a filter which passes frequencies lower than the given cut-off frequency and ignores signals whose frequencies are above the cutoff frequency.

This new image, which $I_1$ represents, constitutes the first level of the Gaussian Pyramid. $I_1$ is created by performing a convolution–like operation wherein the values of pixels in $I_0$ are subjected to a weighted average by a selected window (say a 5 x 5 window), i.e. Gaussian Blurring. Subsequently, image $I_2$ is created by performing the same steps on $I_1$, $I_3$ is created from $I_2$, and up to the $n^{th}$ level.

\begin{figure}[ht]
    \centering
    \includegraphics[width = \linewidth]{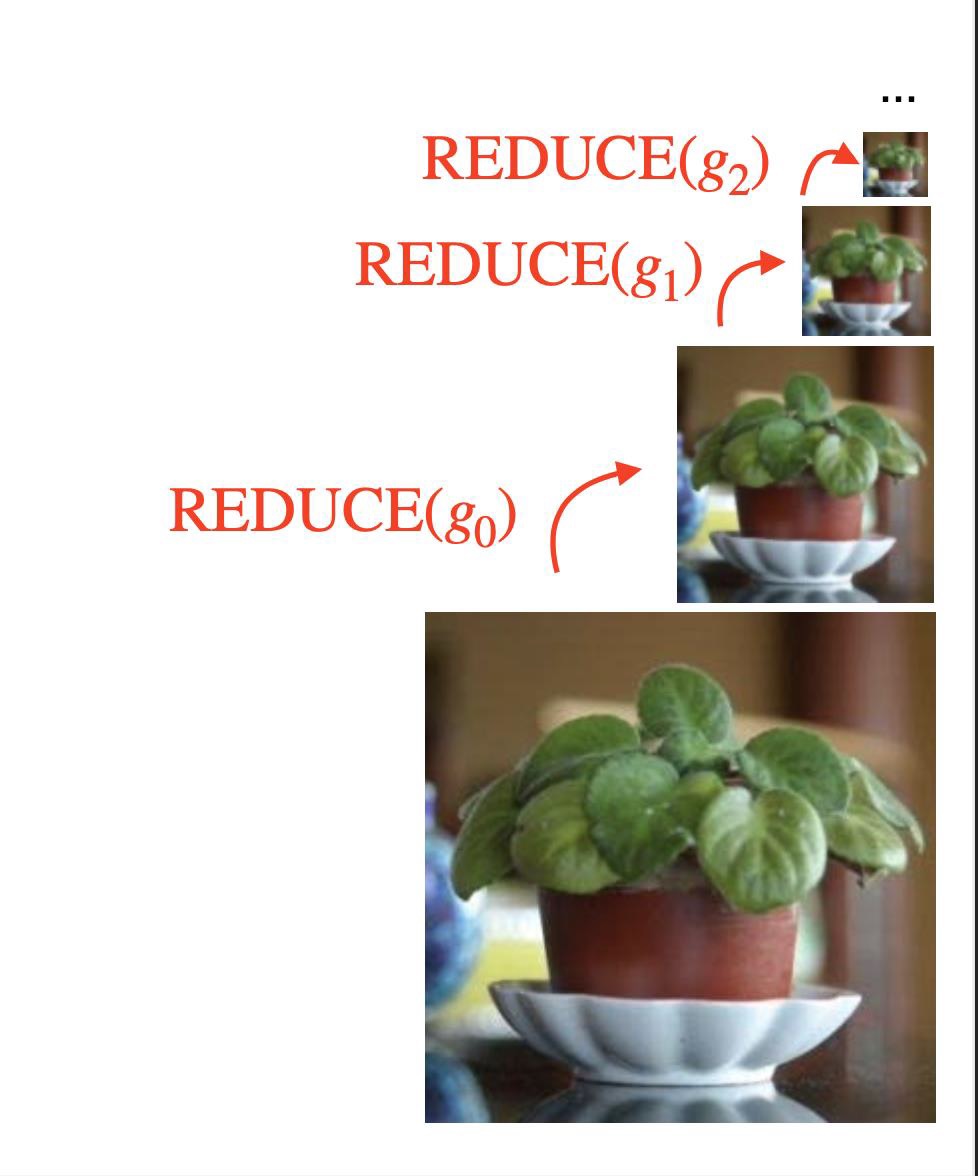}
    \caption{Representation of the REDUCE function. From\\ "https://www.cs.toronto.edu/$\sim$mangas/teaching/320/slides/\\CSC320L10.pdf"}
    \label{Fig4}
\end{figure}

The function performed above is the REDUCE function, wherein:
\begin{equation}
\label{Eq1}
    I_l = REDUCE(I_{l-1})
\end{equation}
which means, for levels $0 < l < N$ and nodes $i$, $j$, $0 < i < C_l$, $0 < j < R_l$:
\begin{equation}
\label{Eq2}
    I_l(i,j) = \sum_{m=-2}^{2}\sum_{n=-2}^{2}w(m,n)I_{l-1}(2i+m,2j+n)
\end{equation}
where $C$ represents the number of columns and $R$ represents the number of rows of $I_0$, and $w$ represents the pattern of weights used to apply the Gaussian blur (see also Fig~\ref{Fig4}).

\begin{figure}[ht]
    \centering
    \includegraphics[width = \linewidth]{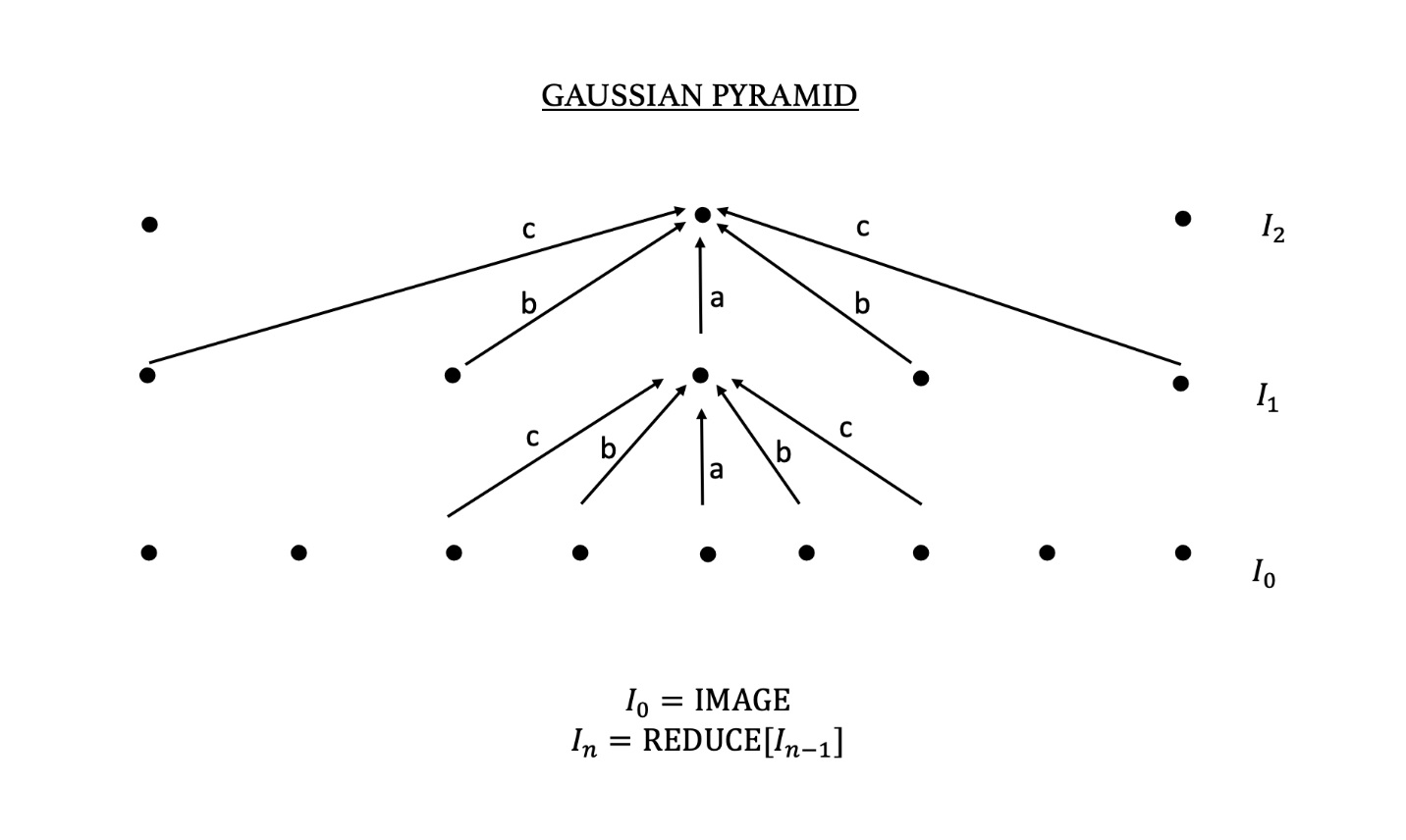}
    \caption{Pictorial Representation of building Gaussian Pyramid where rows $I_0$, $I_1$ and $I_2$ depict the levels of Gaussian pyramid, the block dots represent the nodes, i.e. the multiple pixels which are pooled as we move up the levels of the Pyramid as denoted in Eq.~\ref{Eq2} by the expression $(2i + m, 2j + n)$, and the arrows which represent which pixels cluster up.}
    \label{Fig5}
\end{figure}

Thus, a Gaussian Pyramid is created by applying a Gaussian blur (averaging of pixels using kernels) to an image, followed by reduction of the image size to half of its original dimensions (see Fig~\ref{Fig5}).

\begin{figure}[ht]
    \centering
    \includegraphics[width = \linewidth]{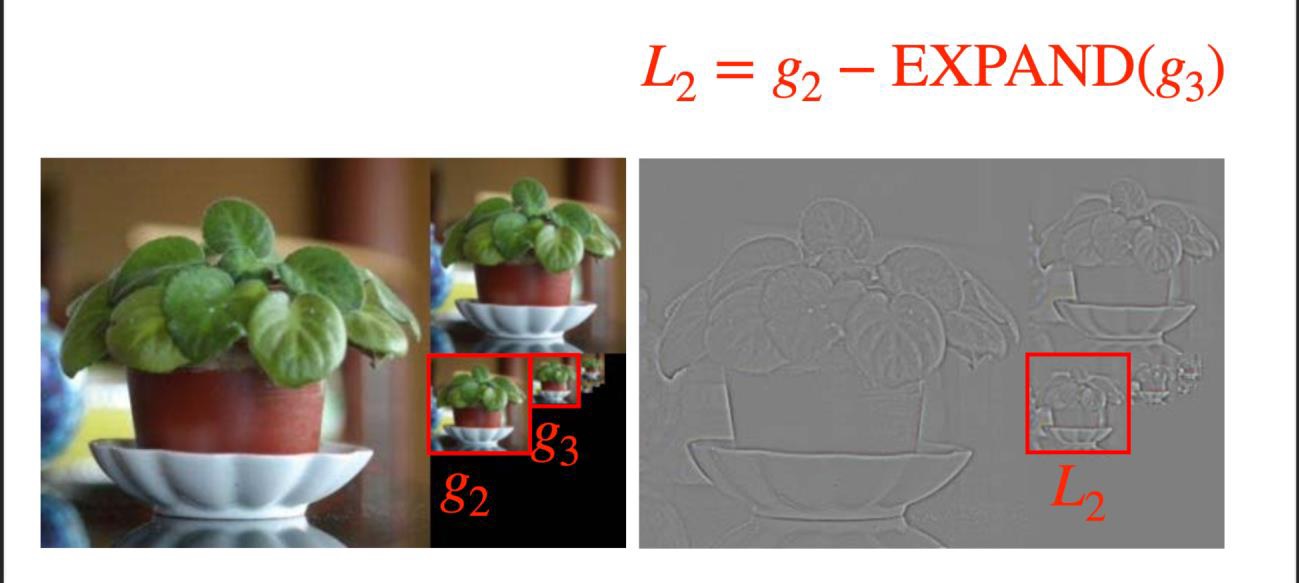}
    \caption{Creation of the Laplacian Pyramid\\ "https://www.cs.toronto.edu/$\sim$mangas/teaching/320/slides/\\CSC320L10.pdf"}
    \label{Fig6}
\end{figure}

\paragraph{Laplacian Pyramid}
The Laplacian Pyramid is created from the Gaussian Pyramid using the EXPAND function as:
\begin{equation}
\label{Eq3}
    L_l = I_l - EXPAND(I_{l+1})
\end{equation}
where $Ll$ stores the difference between the larger and smaller images, i.e, whatever has been blurred by the $REDUCE$ function (Eqs 1, 2) where
\begin{equation}
\label{Eq4}
    EXPAND(I_n) = 4\sum_{m=-2}^{2}\sum_{n=-2}^{2}w(m,n)I_l(\frac{i-m}{2},\frac{j-n}{2})
\end{equation}
Thus, the function enlarges the image from 
$(2^{N-l} + 1) \times (2^{N-l} - 1)$ to $(2^{N-l+1} + 1) \times (2^{N-l+1} - 1)$, i.e., the image is doubled (see Fig~\ref{Fig6}).

The construction of this Laplacian pyramid leads to the down-sampling and the decomposition of the video into spatial frequency bands. These spatial frequency bands are subjected to temporal processing to magnify the motions of our interest. The time series corresponding to a pixel’s value in a given spatial frequency band is considered. An appropriate bandpass filter (a filter which takes in all frequencies and passes the frequencies within a specified range and rejects those outside this given range) is used to extract the frequencies of interest. For example, if we consider the frequency of human respiration which is approximately 0.2 Hz, we can specify a range of 0 - 0.5 Hz to the bandpass filter to extract this frequency.

The next subsection details the temporal processing to magnify the motions of interest.

\subsubsection{Temporal Processing}
\label{Sec3_1_2}
The temporal processing uses the extracted frequencies from the spatial frequency band in section 4.1.1 to look for oscillations of brightness at a given pixel across the frames of the video and magnify them. Since the detailed temporal processing for a complex data is extremely complicated, the temporal processing for a single sinusoidal curve is given below.

Let $I(x, t)$ denote image intensity at time $t$ and position $x$. Due to the translational motion of the image, the expression for intensity with respect to displacement function $\delta (t)$ is modified as:
\begin{equation}
\label{Eq5}
    I(x,t) = f(x+\delta (t))
\end{equation}
Next, we use first-order Taylor Series expansion to approximate about $x$ as:
\begin{equation}
\label{Eq6}
    I(x,t) \approx f(x) + \delta (t) \frac{\partial f(x)}{\partial x}
\end{equation}

Consider $B(x,t)$, obtained by applying a bandpass filter over $I(x,t)$. This filter picks up on everything except for $f(x)$. It is assumed that $\delta (t)$ is also within the allowable band of the filter. We see that:
\begin{equation}
\label{Eq7}
    B(x,t) = \delta (t) \frac{\partial f(x)}{\partial x} 
\end{equation}
Finally, magnification factor $\alpha$ is applied to $B(x,t)$ and added back to $I(x, t)$, which results in
\begin{equation}
\label{Eq8}
    \hat{I}(x,t) = I(x,t) + \alpha B(x,t)
\end{equation}
By combining Equations~\ref{Eq5},~\ref{Eq6}~and~\ref{Eq8} we arrive at:
\begin{equation}
\label{Eq9}
    \hat{I}(x,t) = f(x)+(\alpha + 1)\delta (t)\frac{\partial f(x)}{\partial x}
\end{equation}
This final perturbation is subjected to first-order Taylor Series expansion to obtain the final processed signal as:
\begin{equation}
\label{Eq10}
    \hat{I}(x,t) = f(x+(\alpha+1)\delta (t))
\end{equation}
This final result represents the spatial displacement $\delta (t)$ of image $f(x)$ at time $t$ which is magnified by order of $(\alpha + 1)$, showing that this processing results in magnified motions.

The entire Eulerian Video Magnification method is summarized in Fig~\ref{Fig7} below.

\begin{figure}[ht]
    \centering
    \includegraphics[width = \linewidth]{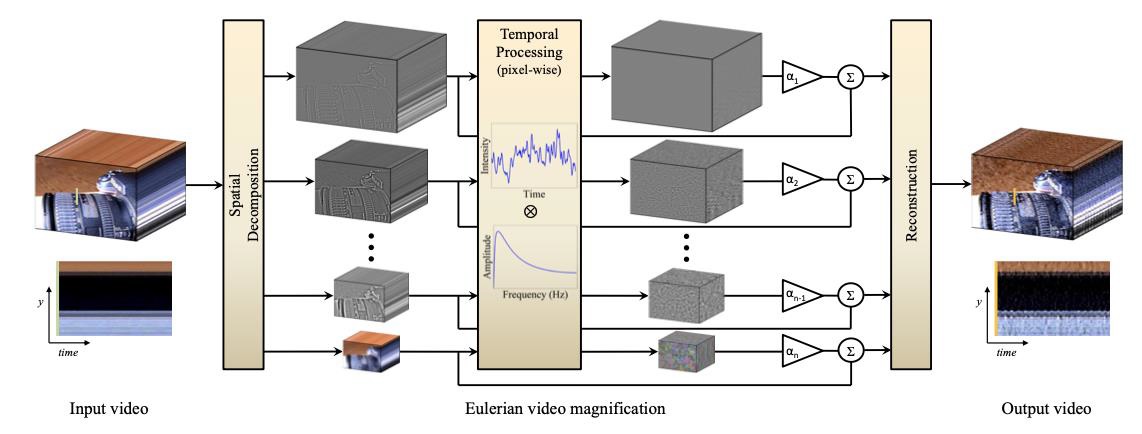}
    \caption{Pictorial Representation of Eulerian Video Magnification pipeline where the input video is first decomposed to the input video into spatial frequency bands using a temporal filter. The decomposed bands are then amplified by magnification factor $\alpha$ and added back to the original signal, thus magnifying motions of interest. These are finally collapsed to generate the output video~\cite{Wu}.}
    \label{Fig7}
\end{figure}

\subsection{Skeletonization}
\label{Sec3_2}
The Skeletonization algorithm by Vikas Gupta~\cite{Gupta} has been used to detect key points in the hand to help locate micro-motions in the hand. This algorithm proposed by Gupta is based on the Hand Key-point Detection in Single Images using Multiview Bootstrapping by Simon et al.~\cite{Simon} and Convolutional Pose Machines by Wei et al~\cite{Wei}. This section details the method described in these papers and the particulars of Gupta’s method.

\begin{figure}[ht]
    \centering
    \includegraphics[width = \linewidth]{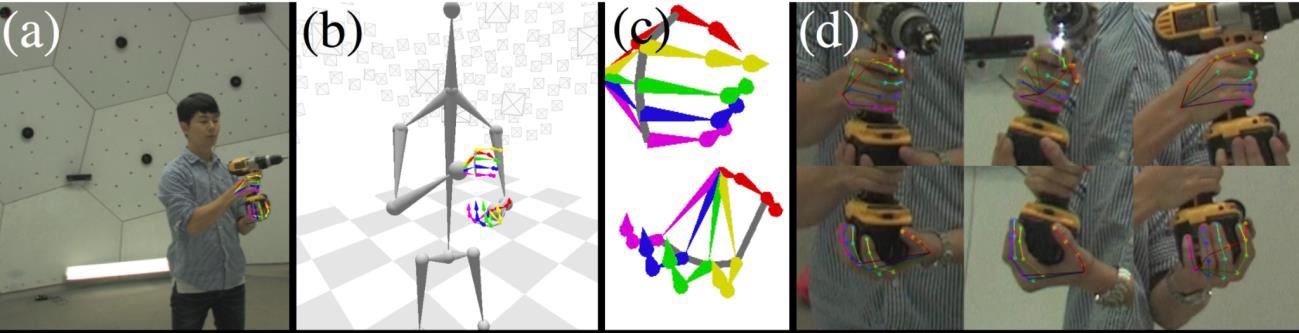}
    \caption{Working of Multi View Bootstrapping~\cite{Simon}.}
    \label{Fig8}
\end{figure}

Simon et al.~\cite{Simon} detail a method to use small datasets which are annotated to detect key points in the “best views” and filter out wrong detections using 3D triangulation. This method aims to better key point detection of the hand and minimize incorrect detection due to minor or significant occlusions by creating a multi-camera view system and training the detector on these multiple views. This method is powerful as it helps label key points in the hand that are almost impossible to label due to occlusion. This technique in the multi-camera setup allows for reconstructions in scenarios such as orchestras or theatre. This multi-view bootstrapping training of data is started off by using a detector $d()$ which maps an image patch $l \epsilon R^{w \times h \times 3}$ to $P$ key points, each with detection confidence of $c_p$:
\begin{equation}
\label{Eq11}
    d(I) \rightarrow \{(x_p,c_p) \mbox{ for } p \in [1 \cdots P] \}
\end{equation}
Each of these points $p$ corresponds to a different feature of the hand. The assumption made is that there is only a single view of each $p$ in selected image patch $I$. Detector $d_0$ is trained on images $I_f$, which have a set of labeled key points \{$y^f$\} This training set which is obtained initially is denoted by $T_0$,which has $N_0$ elements (see Fig~\ref{Fig8} for an illustration of this method).

Thus,
\begin{equation}
\label{Eq12}
    T_0 := (I^f,\{y^f\} \mbox{ for } f \in [1 \cdots N_0])
\end{equation}
Given an initial detector $d_0$, and an unlabeled dataset of multi-view images, the detector must generate a set of labeled images depicted by $T_1$. $T_1$ is then used to improve the detector using data from $T_0$ and $T_1$. This performance enhancement of detector $T_0$ is done using a process known as 3D triangulation.

The main aim of 3D triangulation is to correct the faulty detections on the first trial of a given view by using detections from other unlabeled training data and creating a correct 2D projection of the key-point without occlusion.

Let us consider a frame of a video that has $V$ views. The detector $d_i$ is trained on every image $I_v^f$, which results in a set of 2D key points detected denoted by $\boldsymbol{D}$
\begin{equation}
\label{Eq13}
    \boldsymbol{D} \leftarrow \{ d_i I_v^f \mbox{ for } v \in [1 \cdot V] \}
\end{equation}
As seen previously, for p detected keypoints in each of these frames, there exists a set \{$x_v^f,c_v^f$\} where $x$ denotes the location of the detected key point and $c$ denotes the confidence measure. For robust 3D triangulation, Random Sample Consensus (RANSAC) is used. RANSAC is a method of estimation used on some data with outliers where the outliers do not affect the estimation and prediction of the mathematical model. RANSAC is used on $\boldsymbol{D}$ (Eq 14). Additionally, a confidence threshold of $\lambda$ is applied.
Thus, we obtain the final triangulated points as:
\begin{equation}
\label{Eq14}
    X_p^f = \mbox( argmin ) \coprod P_v(X) - x_p^v \coprod_{2}^{2}
\end{equation}
where $X_p^f \in R^3$ is the 3D triangulated keypoint $p$ in frame $f$, and $P_v(X) \in R^2$ denotes the projection of 3D point $X$ into view $v$.

This method of 3D triangulation is robust. Even though it constructs a lower number of triangulated key points, it makes sure that the entire hand’s detections are accurate to prevent incorrect data for training the subsequent detectors.

\begin{figure}[ht]
    \centering
    \includegraphics[width = \linewidth]{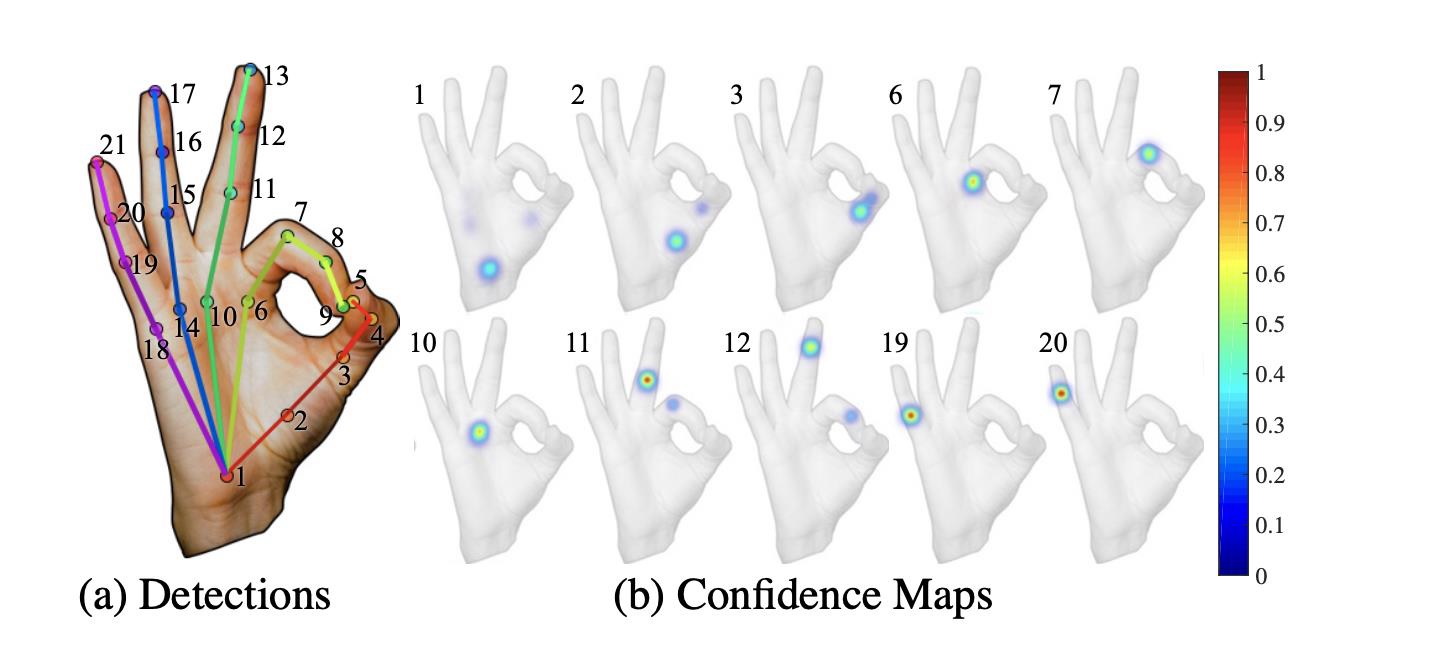}
    \caption{Confidence maps produced in the Multiview Bootstrapping~\cite{Simon}.}
    \label{Fig9}
\end{figure}

These detected triangulated keypoints are now scored to detect the best key points. This is to avoid erroneous data in subsequent training sets. This scoring of the frame is done by summating the detection confidences produced. The “best” frame is picked by observing which frame has the highest confidence summation (see Fig~\ref{Fig9}).
\begin{equation}
\label{Eq15}
    \mbox{score}(\{X_p^f\}) = \sum_{p \in [1 \cdots P]}\sum_{v \in I_p^f} c_p^v
\end{equation}
Finally, the next detector, i.e., the $d_{i+1}$ is constructed using the $N$ best projections. The N best projections training set is determined as:
\begin{equation}
\label{Eq16}
    \begin{aligned}
        T_{i+1} := \{(I_v^{s_n},\{P_v(V_v^{s_n}): v \in [1 \cdots V], p \in [1 \cdots P]\} \\
        \mbox{ for } n \in [1 \cdots N]\}
    \end{aligned}
\end{equation}
where $P_v(V_v^{s_n})$ depicts the projection of point $p$ for frame index $S_n$ into a view $v$.

The detectors $d_i$ used in this method are Convolutional Pose Machine (CPM) detectors. Convolutional Pose Machines (CPMs) [12] implement a method that is used to estimate human poses. CPMs consist of multiple convolutional neural networks which produce belief maps. The previous layer’s belief map and image features are fed in as input to the subsequent layer for each layer in the CPM as depicted in Fig 10 below.

\begin{figure}[ht]
    \centering
    \includegraphics[width = \linewidth]{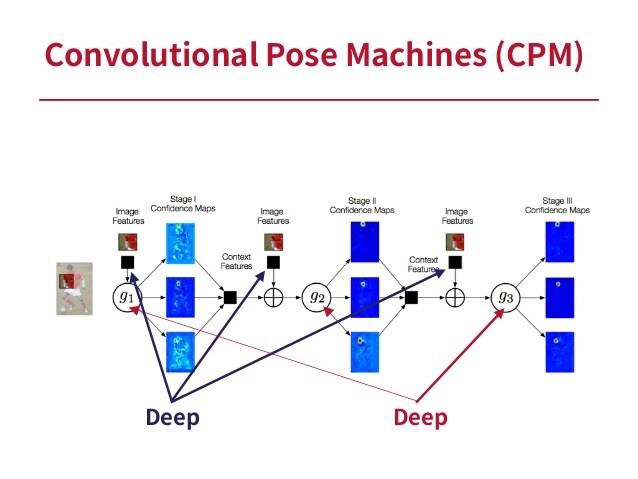}
    \caption{Architecture of Convolutional Pose Machine.}
    \label{Fig10}
\end{figure}

The highlights of the CPM model are:
\begin{enumerate}
    \item Its learning process enables it to learn implicit spatial models using a sequence-of-convolutional-networks architecture.
    \item The inclusion of intervening or intermediate supervision, i.e., the supervision of learning in intermediate layers of very deep networks to prevent the vanishing of gradients. The vanishing of gradients is the diminishing of back-propagated gradients due to propagation through multiple layers of the network (especially very deep networks). Although this intermediate supervision has been used in training in other related works for classification problems, CPMs have tried to incorporate this feature for a structured task such as pose estimation.
    \item Training the architecture as mentioned earlier to learn and understand image dependent spatial models and the critical features of the image.
    \item Testing the model on standard datasets such as MPII, LSP, and FLIC datasets and achieving innovative and new results.
\end{enumerate}
In the method developed by Simon et al.~\cite{Simon}, the convolutional stages of the predefined VGG-19 network are used.

Vikas Gupta uses the above method in~\cite{Gupta} to help detect 22 3D keypoints of the hand, representing the background and the other 21 representing the hand. The frame is first converted to a blob and creates probability maps as matrices to estimate the locations of the key points. Based on a threshold of confidence (such as $\lambda$ as mentioned above), the point is either selected or rejected. After selecting the appropriate key points, the algorithm draws circles to show the key points on the frame. The algorithm also draws lines to connect the key points.

This approach was used through a package from Gupta’s method for this paper. The CPM and the skeletonization algorithms were pre-trained into a model by Gupta and made available~\cite{Gupta}. They were not trained on the raw data acquired for this paper.

\subsection{K Nearest Neighbor (kNN) Classifier}
\label{Sec3_3}
The kNN classifier is a non - parametric machine learning algorithm. As a non-parametric algorithm, it does not make strong assumptions about the form of the mapping function. Due to this functionality, it can learn any functional form from the data provided to it. The kNN algorithm works by correlating a data point with its $k$ nearest similar neighbors and classifying it. During the classification of a data point, the Euclidean distances between it and neighboring points are calculated.

\begin{figure}[ht]
    \centering
    \includegraphics[width = \linewidth]{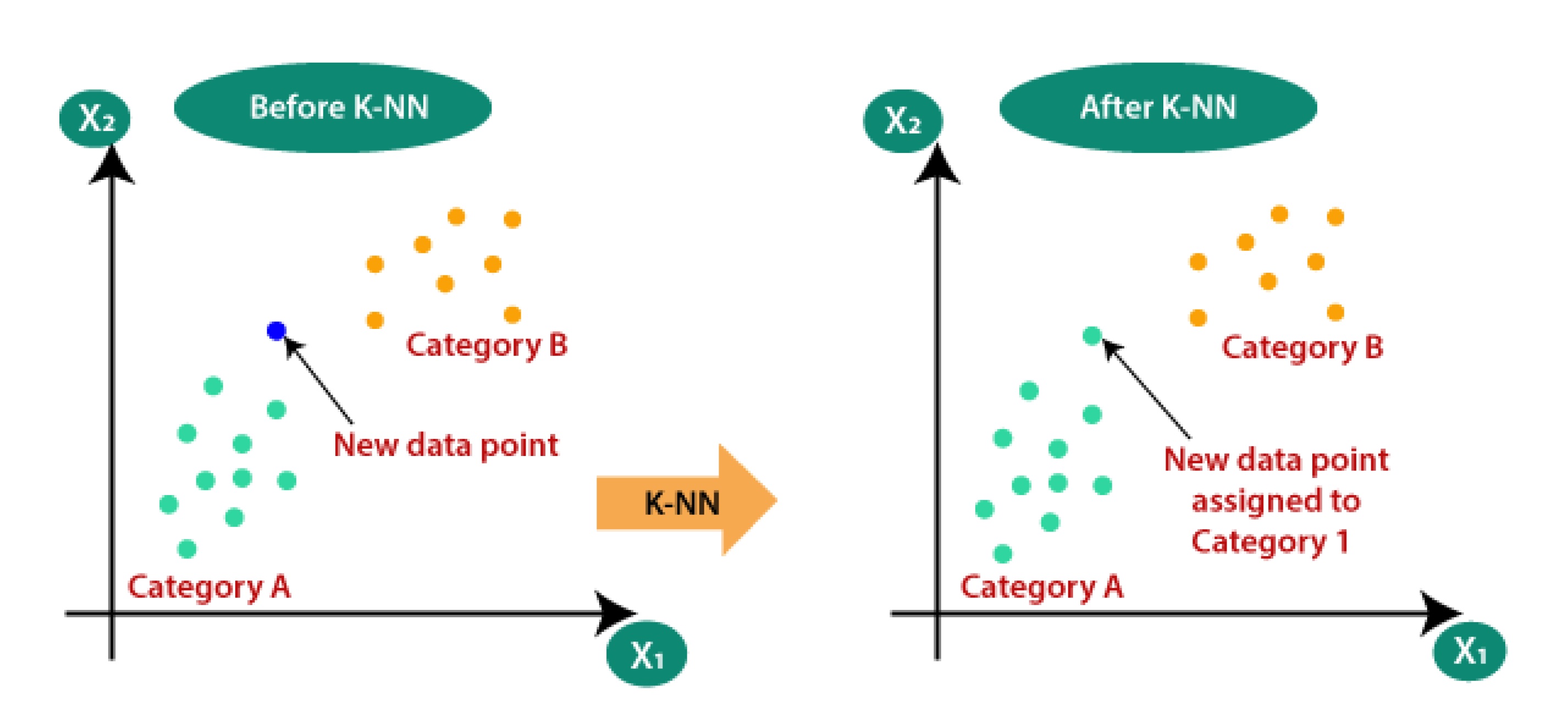}
    \caption{Depiction of the working of kNN algorithm.}
    \label{Fig11}
\end{figure}

The Euclidean distance is calculated by:
\begin{equation}
\label{Eq17}
    d(x,y) = \sqrt{\sum_{i=1}^n (y_i-x_i)^2}
\end{equation}
Based on the values of $k$ and the minimum Euclidean distances calculated, the point is classified as depicted in the Fig~\ref{Fig11}.

\section{Method}
\label{Sec4}
\begin{figure}[ht]
    \centering
    \includegraphics[width = \linewidth]{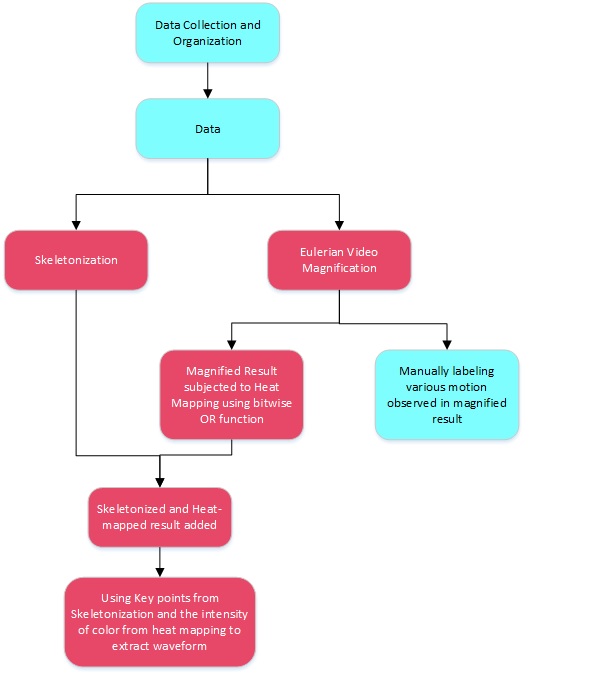}
    \caption{Image representing the pre - processing pipeline of the proposed method. The pink boxes depict computerized components of pre-processing and the blue boxes depict the manual components.}
    \label{Fig12}
\end{figure}

This work proposes a method to detect and classify micro-motions in the hand using Waveform Analysis present in the hand using Eulerian Video Magnification, Skeletonization, Heat - Mapping, and the kNN machine learning algorithm. The videos are first pre-processed to gain relevant information about the micro movements in the hand. This pre-processing to obtain the final waveform is shown above in Fig 12.

\subsection{Data Acquisition}
\label{Sec4_1}
Smartphone videos of both hands of 45 subjects, i.e. 90 datasets, were collected. Due to COVID - 19 restrictions, subjects were asked to self - record videos from their homes. The gender ratio of subjects taken for this test is 1:1. All subjects considered in this experiment are Indian. The mean age of all subjects is 51 years, with the youngest being 13 and the oldest being 90 years old.

Data were acquired by asking subjects to record videos on smartphones in lit conditions. The videos were taken by keeping the hand steadily on the arm of a chair, facing down, and keeping the camera on a stand at one meter from the subject as depicted in Figs 13 and 14 below.

\begin{figure}[ht]
    \centering
    \includegraphics[width = \linewidth]{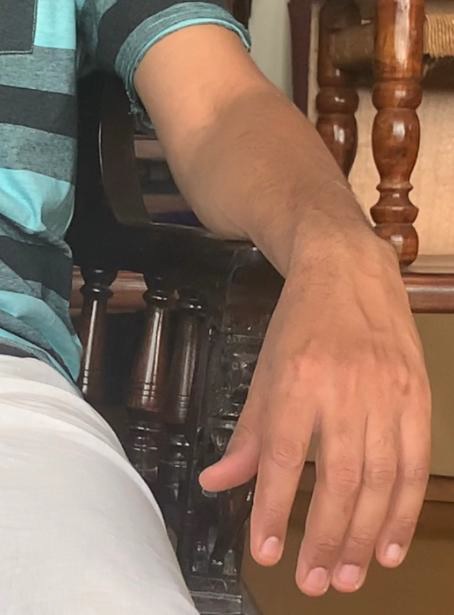}
    \caption{Still from ideal video of Subject 001.}
    \label{Fig13}
\end{figure}

\begin{figure}[ht]
    \centering
    \includegraphics[width = \linewidth]{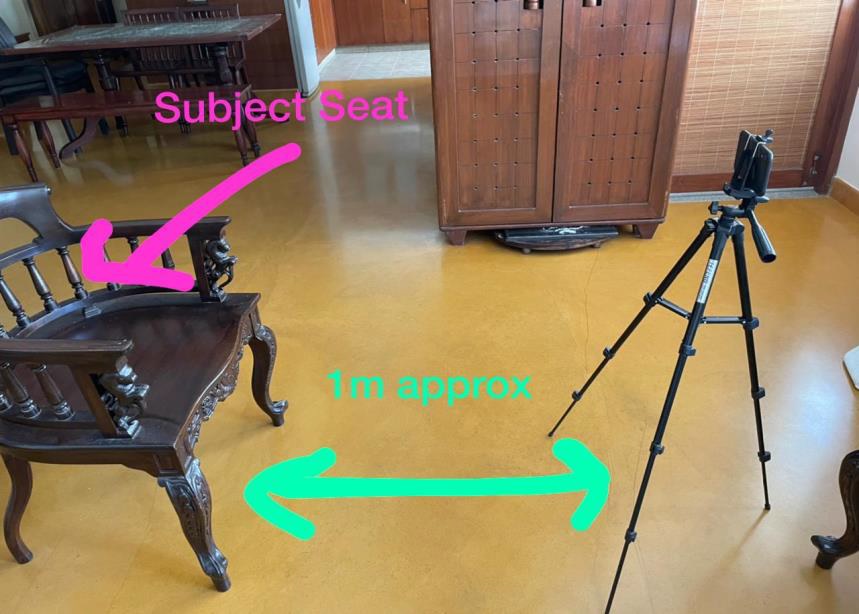}
    \caption{Depiction of set-up for video data acquisition.}
    \label{Fig14}
\end{figure}

In specific videos, the original videos were not of the appropriate size for creating the three levels of the Laplacian Pyramid, which is built during the Eulerian Video Magnification algorithm. Due to this, 1, 2, or 3 rows or columns of zeros are added to the image (images are padded), thus changing the size. Therefore, during the process of heat-mapping, due to the usage of both the original and magnified video, the original image is resized to the same size as the magnified result.

\subsection{Pre-processing steps for Classification}
\label{Sec4_2}
\begin{figure}[ht]
    \centering
    \begin{subfigure}[b]{0.48\linewidth}
    \includegraphics[width = \linewidth]{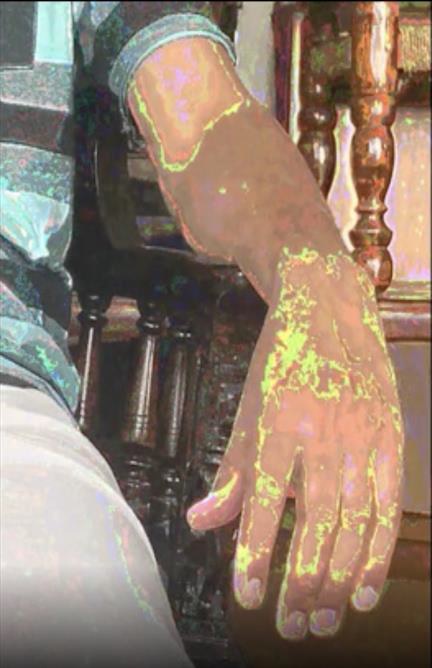}
    \caption{}
    \end{subfigure}
    \begin{subfigure}[b]{0.48\linewidth}
    \includegraphics[width = \linewidth]{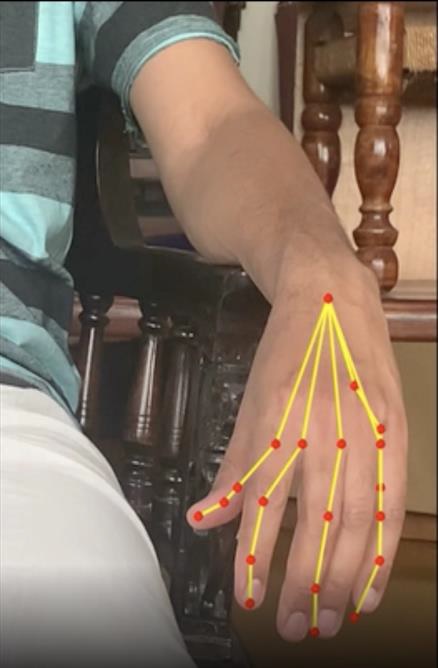}
    \caption{}
    \end{subfigure}
    \caption{(a) Heatmap and (b) Skeletonization.}
    \label{Fig15}
\end{figure}

The videos acquired directly from subjects were first subjected to skeletonization which resulted in the detection of 22 key points in the hand. The key points are essential to mark the location of the micro-motions. Additionally, the skeletonization algorithm also joins these key points to form a skeleton-like structure on the hand, as seen in Fig~\ref{Fig15}.

These lines are not essential to the method discussed here; however, they serve as additional landmarks to help locate the motions of interest.

The original data is also subjected to Eulerian Video Magnification in parallel. This results in the magnified video, where various micro-motions in the hand such as the pulse, tremor components, random movements in the hand, etc., become visible to the human eye.

This magnified result is then heat-mapped as seen in Fig. 15(b). Since available heat-mapping algorithms are used to magnify larger motions, they do not perform effectively for this use case. Thus, here a bit-wise OR operator was used between the original video and the magnified video which yields a suitable heatmap that can pick up on the subtle changes in position, i.e., the small motions of the various parts of the hand.

\begin{figure}[ht]
    \centering
    \includegraphics[width = \linewidth]{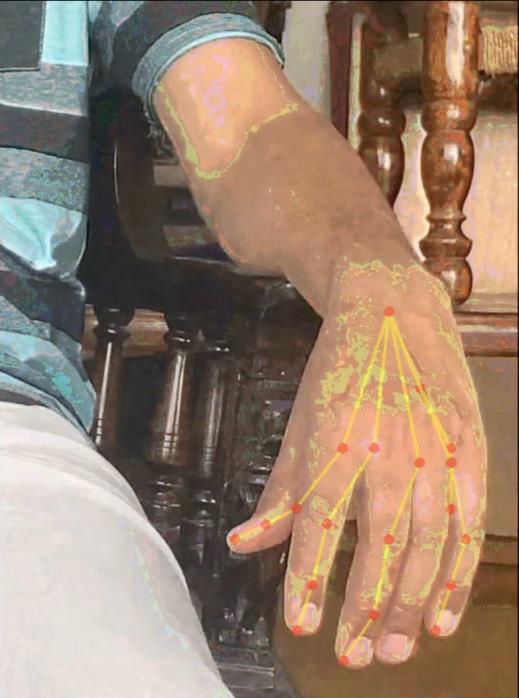}
    \caption{Overlap of Skeletonized result and Heatmap.}
    \label{Fig16}
\end{figure}

Finally, the results from the heat-mapped and skeletonized results are averaged to overlap the skeletonization and heat-mapped results (see Fig~\ref{Fig16}). This overlapped video is used to label motion. The intensity of color and key points detected in the hand are essential features for classifying the motions in different parts of the hand.

\subsection{Labelling Data}
\label{Sec4_3}
Two stages of data labelling were followed for the Machine Learning Classification task.

\subsubsection{Stage 1 - Labelling motions occurring in original video submitted by the subject}
\label{Sec4_3_1}
In this first stage, only the original video was considered. Extra background motions such as the movement of people walking in the background, the movement of a swing, etc., and the movement of the entire frame was labeled as background motion.

\subsubsection{Stage 2 - Labelling motion from Motion-magnified result}
\label{Sec4_3_2}
\begin{figure}[ht]
    \centering
    \begin{subfigure}[b]{0.48\linewidth}
    \includegraphics[width = \linewidth]{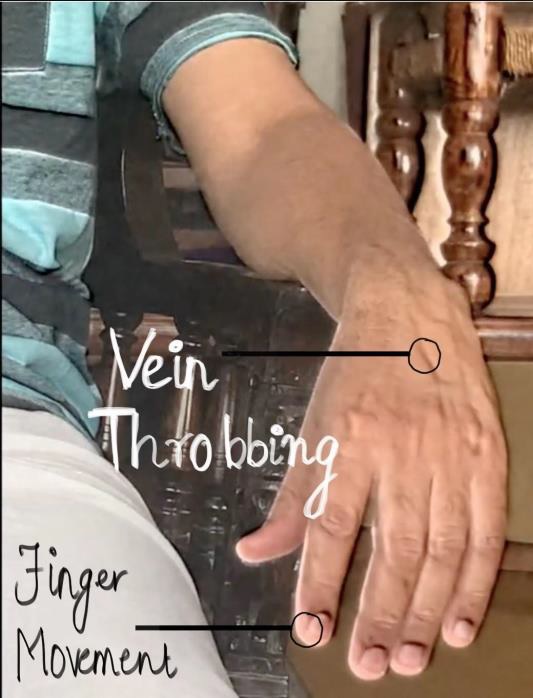}
    \caption{}
    \end{subfigure}
    \begin{subfigure}[b]{0.48\linewidth}
    \includegraphics[width = \linewidth]{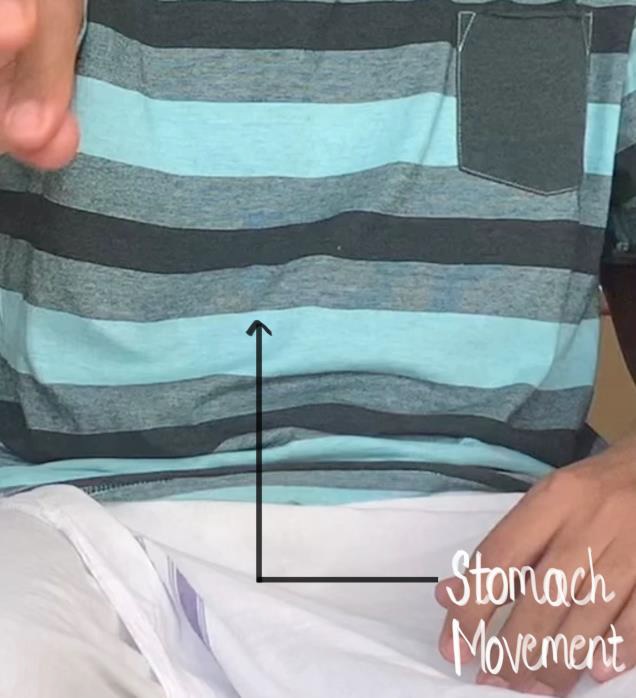}
    \caption{}
    \end{subfigure}
    \caption{Stills from magnified videos depicting (a) Vein throbbing and Finger movement, (b) movement of the stomach.}
    \label{Fig17}
\end{figure}

In this stage, the magnified result is used to label certain other motions. These are finger, hand movements or tremors and throbbing of blood vessels in hand. In specific videos where subjects recorded their entire middle bodies and hands, movement of the stomach during breathing was also labeled. These labels are detailed in the Fig.~\ref{Fig17}.

The waveforms above (see Fig~\ref{Fig17}) are created using the RGB values of select pixels in the right stomach of subject 001. These pixels are selected using the pre-processing steps detailed in the previous subsection~\ref{Sec4_2}. Within the hand, pixels are picked using the results of the heatmap and skeletonization. Pixels where all pre-processing steps match are selected.

\subsection{Waveform Creation}
\label{Sec4_4}
\begin{figure}[ht]
    \centering
    \begin{subfigure}[b]{\linewidth}
    \includegraphics[width = \linewidth]{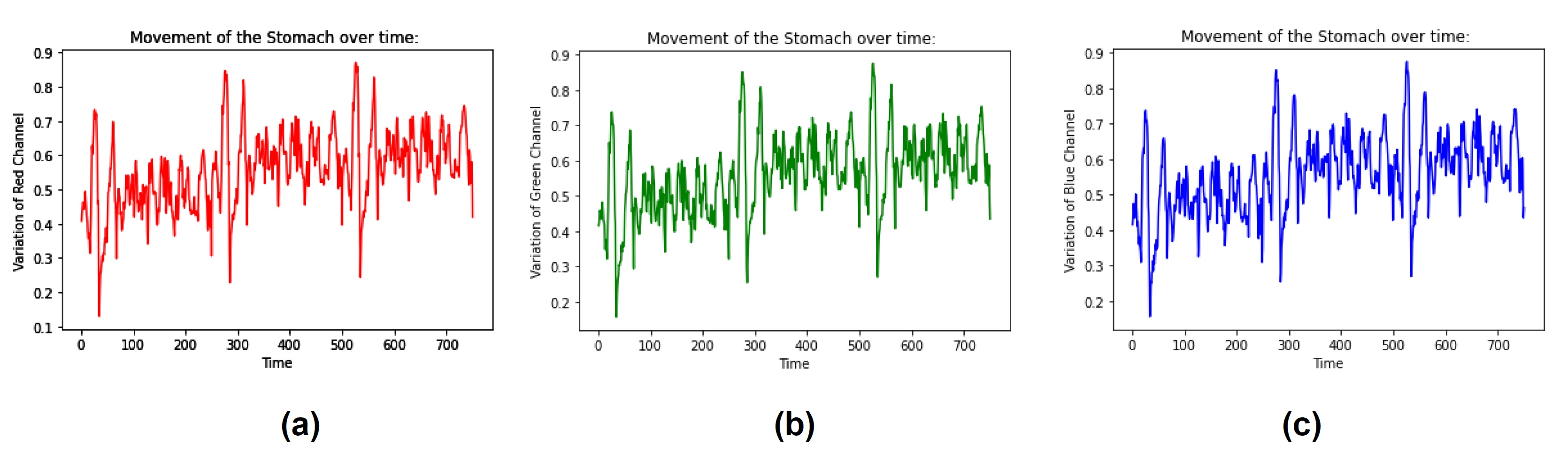}
    \end{subfigure}
    \begin{subfigure}[b]{0.35\linewidth}
    \includegraphics[width = \linewidth]{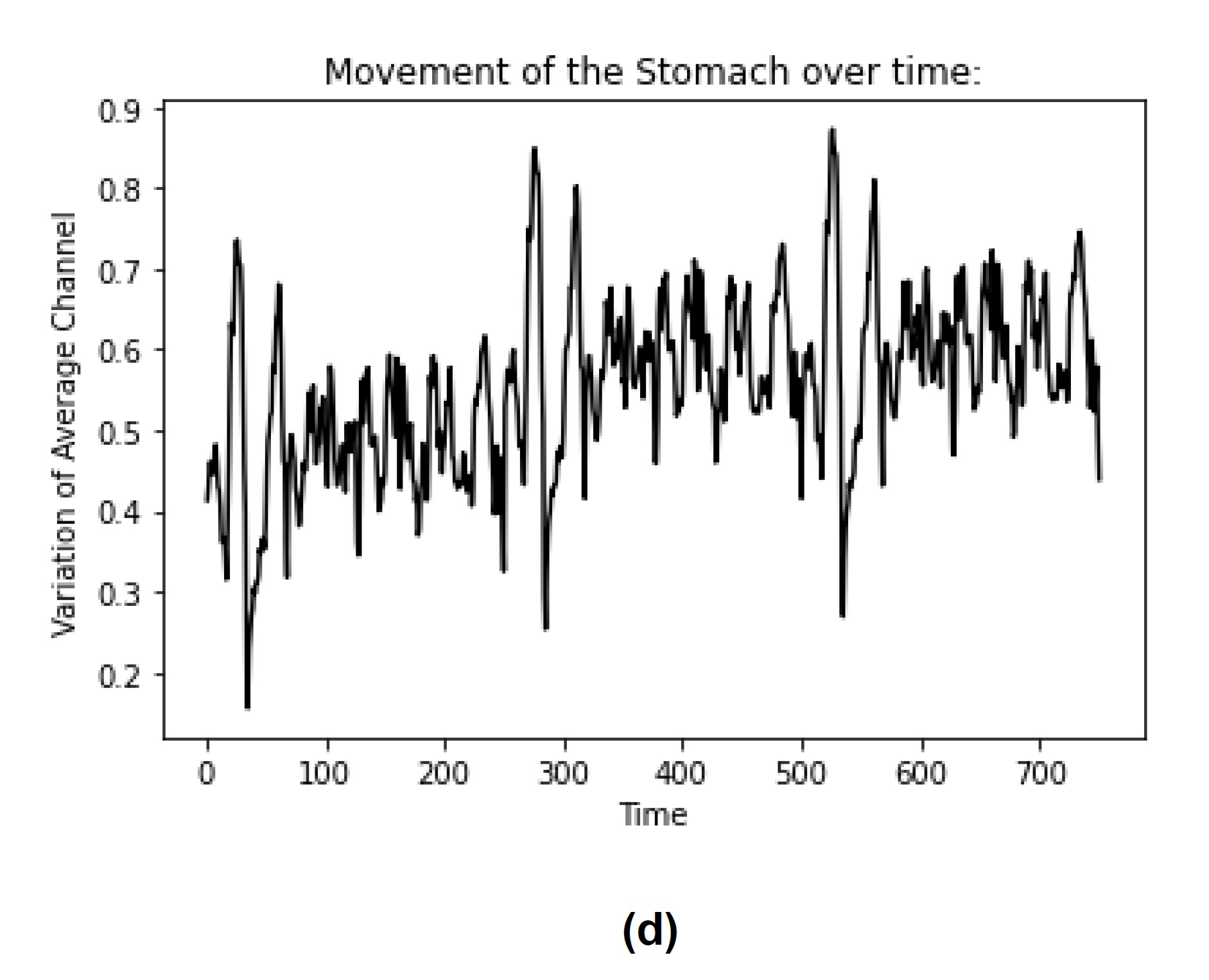}
    \end{subfigure}
    \caption{Waveforms depicting the movement of the stomach due to respiration.}
    \label{Fig18}
\end{figure}

\begin{figure}[ht]
    \centering
    \includegraphics[width = \linewidth]{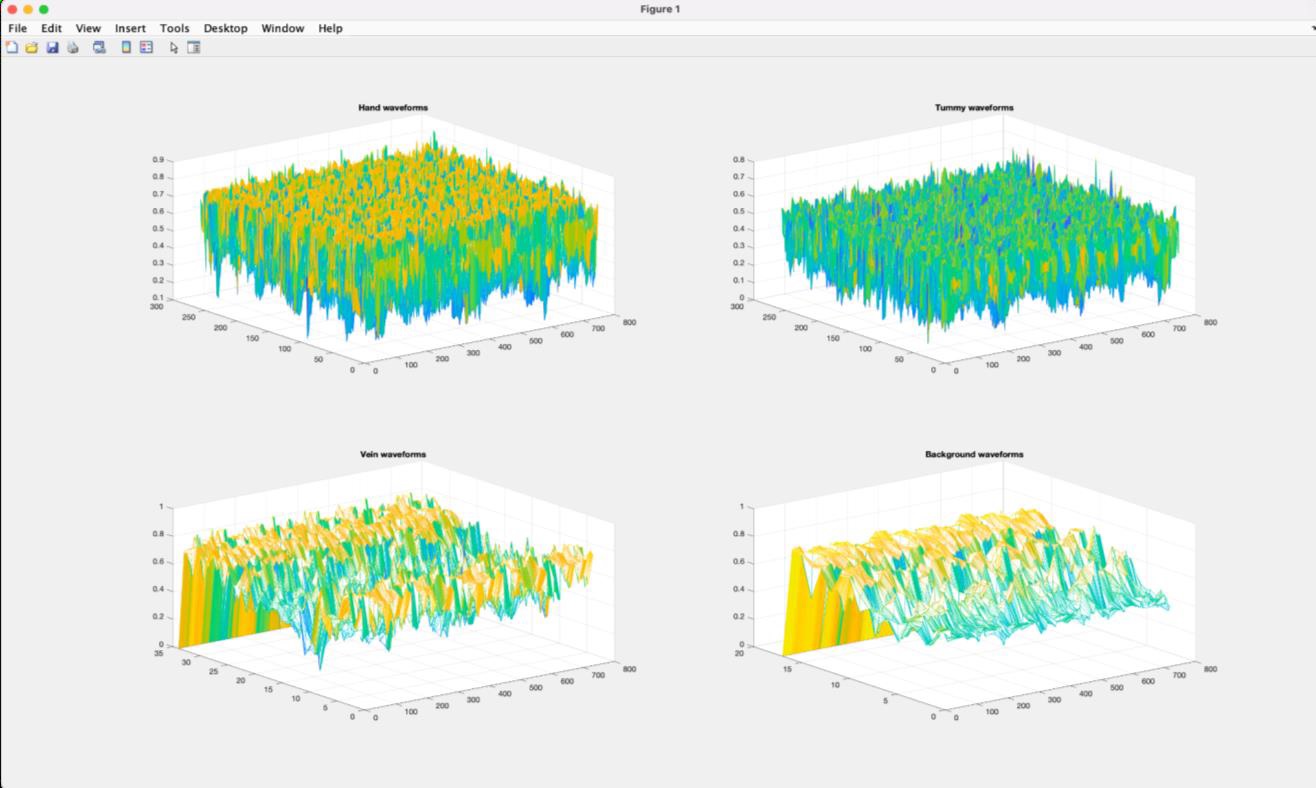}
    \caption{Mesh plot created from waveforms of all pixels of Subject 010.}
    \label{Fig19}
\end{figure}

For selection of pixels outside the hand, only heatmap and manual markings are used. This is how the points for background motion and stomach motion are found. Initially, all three-color channels’ waveforms were plotted (Fig~\ref{Fig18}~(a-c)). However, it was observed that since color magnification did not take place across time, the average of the three channel waveforms could be taken as the final waveform for the point (Fig~\ref{Fig18}(d)). Using this method, waveforms are synthesized for every selected pixel from every video 
(refer to Fig.~\ref{Fig19} below which depicts the waveforms of all pixels of a particular video).

\subsection{Classification}
\label{Sec4_5}
The labelled waveforms obtained were finally passed into a kNN model for classification. The waveforms were limited to the central 10 seconds of data from each subject for inputting into the kNN. These waveform inputs into the kNN were vectors of the averaged RGB values of the pixels in each region of each frame, i.e. pixels of each subject under a particular label.

For classification, the kNN model was used. Multiple past works and papers have assessed the usefulness of different machine learning (Bagged Trees, Decision Trees, Weighted and Unweighted kNNs, etc.) and deep learning models (Deep Belief Network, Deep Neural Network and Long-Short Term Memory, and Recurrent Neural Networks, etc.) for the classification of waveforms. Alghamdi et al.~\cite{Alghamdi} detail the use of kNNs for waveform classification and show promising results. Additionally, for this project, the data collected is highly over-represented by the ‘hand motion’. There is almost 3 times the amount of data for the ‘hand motion’ as there is for ‘background’ or ‘vein motion’. Due to this bias in data, kNNs were preferred over Deep Learning methods.

For this project, Google Colab was used to train and test the kNN. The data was split into three random mini - cohorts for training. The cohorts were each trained on a different model. The outputs of the three models were plotted into confusion matrices to assess the performance of the model. Additionally, the accuracy of the models with respect to the value of ‘k’ nearest neighbors was assessed by plotting graphs of the accuracy versus the value of ‘k’. This analysis revealed that the highest accuracy was when $k = 3$.

\section{Results and Analysis}
\label{Sec5}
\subsection{Eulerian Video Magnification}
\label{Sec5_1}
\begin{figure}[ht]
    \centering
    \begin{subfigure}[b]{0.48\linewidth}
    \includegraphics[width = \linewidth]{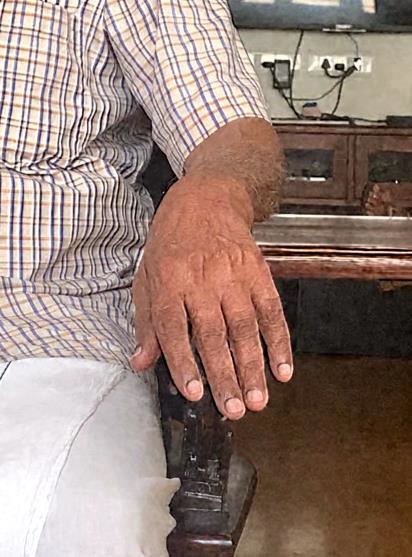}
    \caption{}
    \end{subfigure}
    \begin{subfigure}[b]{0.48\linewidth}
    \includegraphics[width = \linewidth]{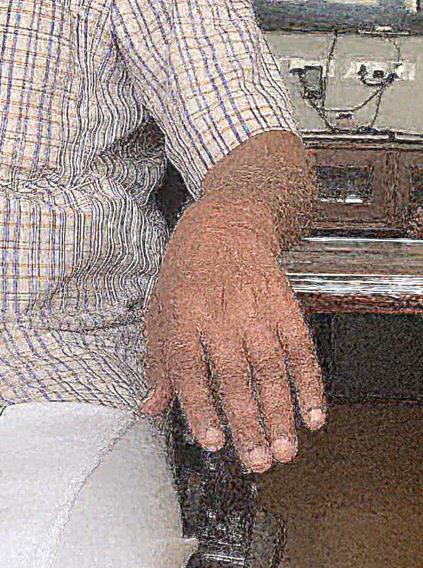}
    \caption{}
    \end{subfigure}
    \caption{Contrast in results between MATLAB and Python implementations where (a) represents the MATLAB and (b) represents the Python implementation.}
    \label{Fig20}
\end{figure}

Two implementations of the Eulerian Video Magnification code were used in this experiment - Python and Matlab. The Python implementation was found on Github~\cite{Zhao} and was written by Flyingzhao. This implementation clarified the code as it was simplified. It magnified the motions properly in all 45 datasets. However, there was an un-explainable "software" noise seen in the videos produced by this method, as seen in Fig.~\ref{Fig20}(b). This noise may be due to an overflow of colors in the pixels. The Matlab code provided by Wu et al.~\cite{Wu} gave expected results as detailed in the original paper “Eulerian video magnification for revealing subtle changes in the world.” This paper provided noiseless video and enabled the visibility of the micro motions in the hand as seen in Fig.~\ref{Fig20}(a). The Eulerian Video Magnification algorithm performed satisfactorily without any visible artifacts or noise in the videos of all different subjects. Subjects had different recording environments such as different lighting settings, different frame sizes, and different cameras and camera angles.

\subsection{Heat-mapping}
\label{Sec5_2}
\begin{figure}[ht]
    \centering
    \begin{subfigure}[b]{0.48\linewidth}
    \includegraphics[width = \linewidth]{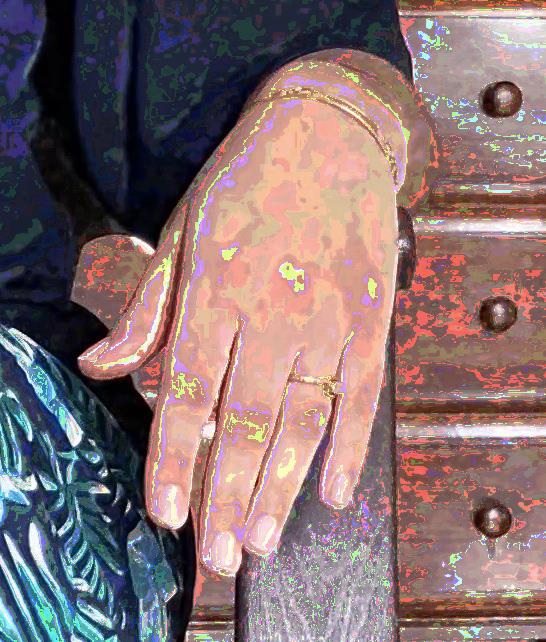}
    \caption{}
    \end{subfigure}
    \begin{subfigure}[b]{0.48\linewidth}
    \includegraphics[width = \linewidth]{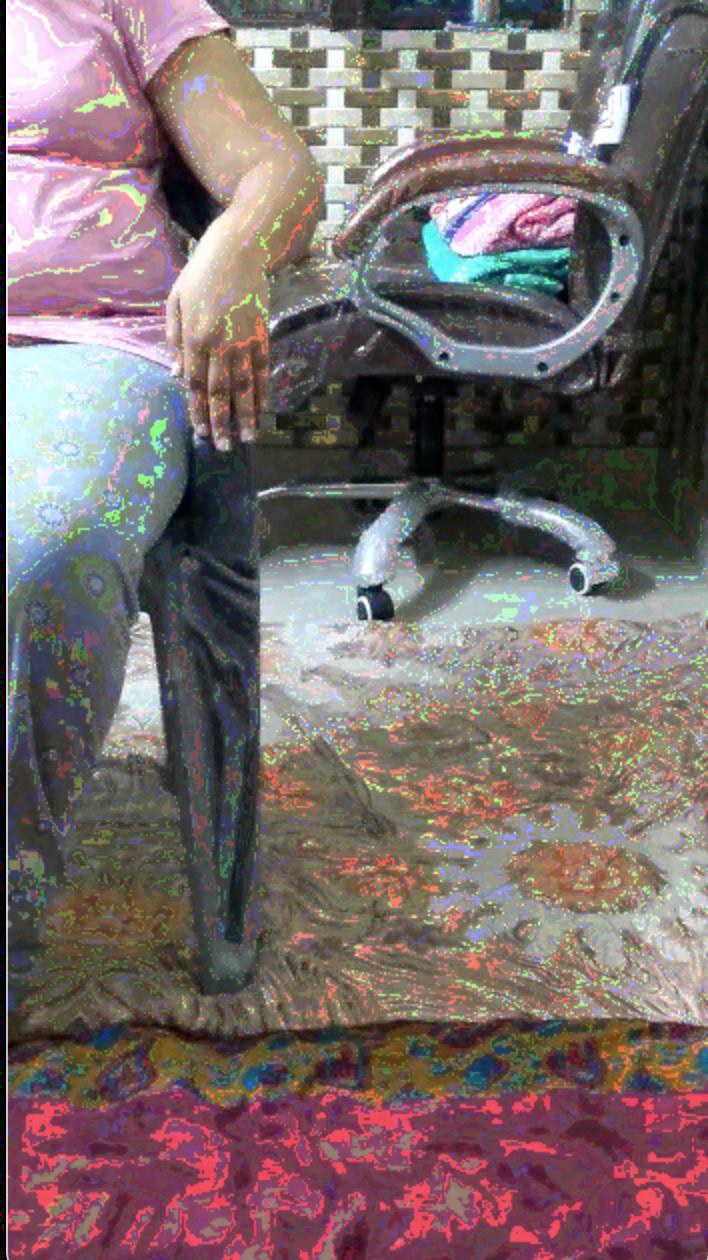}
    \caption{}
    \end{subfigure}
    \caption{Hand close to the camera (a) versus Hand far away from the camera (b).}
    \label{Fig21}
\end{figure}

Heat-mapping was done by performing the Bit-wise OR operation on the original video and the magnified video. This heat mapping highlighted the regions in the hand which had suffered motion after motion magnification. The analysis of results was done using various parameters. One was the closeness of the camera to the hand. It was observed that none of the videos which were far away from the camera yielded useful results, and thus, a correlation can be set up between the distance of the hand from the camera and the quality of the heat-map produced. This is highlighted in Figs~\ref{Fig21}(a) and (b).

\begin{figure}[ht]
    \centering
    \begin{subfigure}[b]{0.48\linewidth}
    \includegraphics[width = \linewidth]{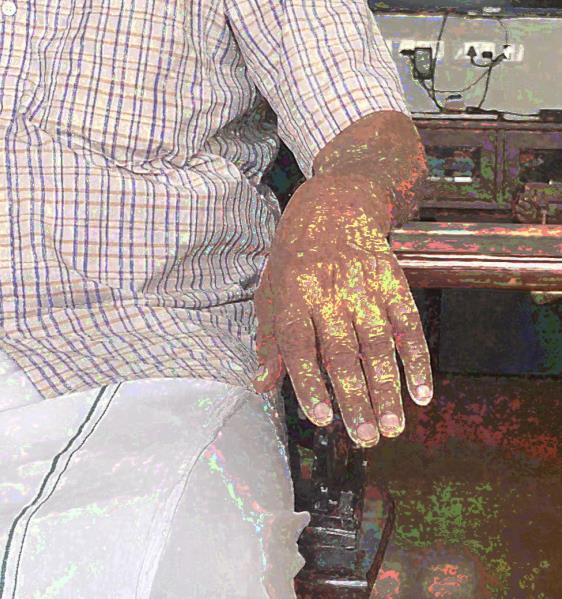}
    \caption{}
    \end{subfigure}
    \begin{subfigure}[b]{0.48\linewidth}
    \includegraphics[width = \linewidth]{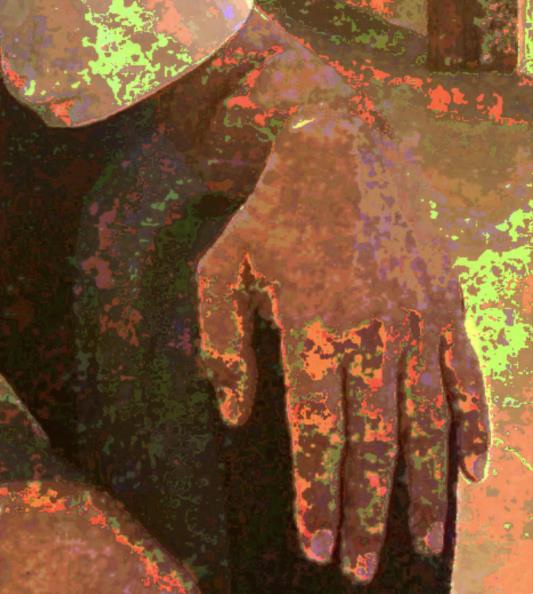}
    \caption{}
    \end{subfigure}
    \caption{Hand in natural lightning (a) versus hand in low lighting (b).}
    \label{Fig22}
\end{figure}

Another parameter used was the difference in the lighting conditions. Specific videos were taken in natural light or well-lit circumstances Fig~\ref{Fig22}(a), while certain videos were taken in low-light or yellow-light conditions Fig~\ref{Fig22}(b). It was observed that the lighting conditions did not affect the quality of the heatmap. It can therefore be concluded that there is no strong correlation between lighting conditions and heat map quality.

The quality of the heat map is determined by the ability of the bit-wise OR operator to detect and highlight the motions produced after magnification. This quality was determined manually.

\subsection{Skeletonization}
\label{Sec5_3}
\begin{figure}[ht]
    \centering
    \begin{subfigure}[b]{0.48\linewidth}
    \includegraphics[width = \linewidth]{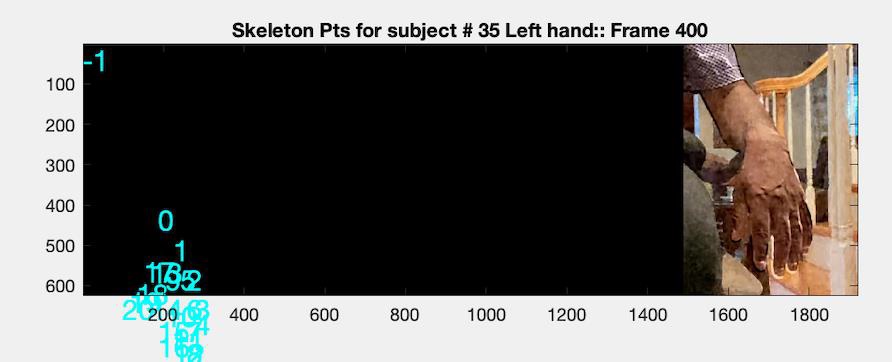}
    \caption{}
    \end{subfigure}
    \begin{subfigure}[b]{0.48\linewidth}
    \includegraphics[width = \linewidth]{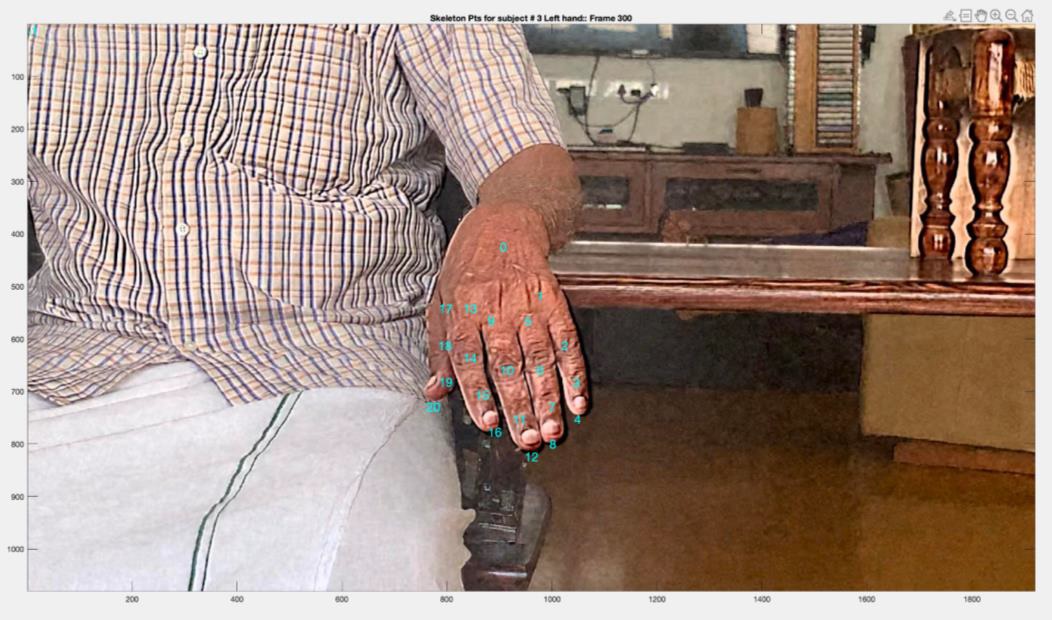}
    \caption{}
    \end{subfigure}    
    \begin{subfigure}[b]{0.48\linewidth}
    \includegraphics[width = \linewidth]{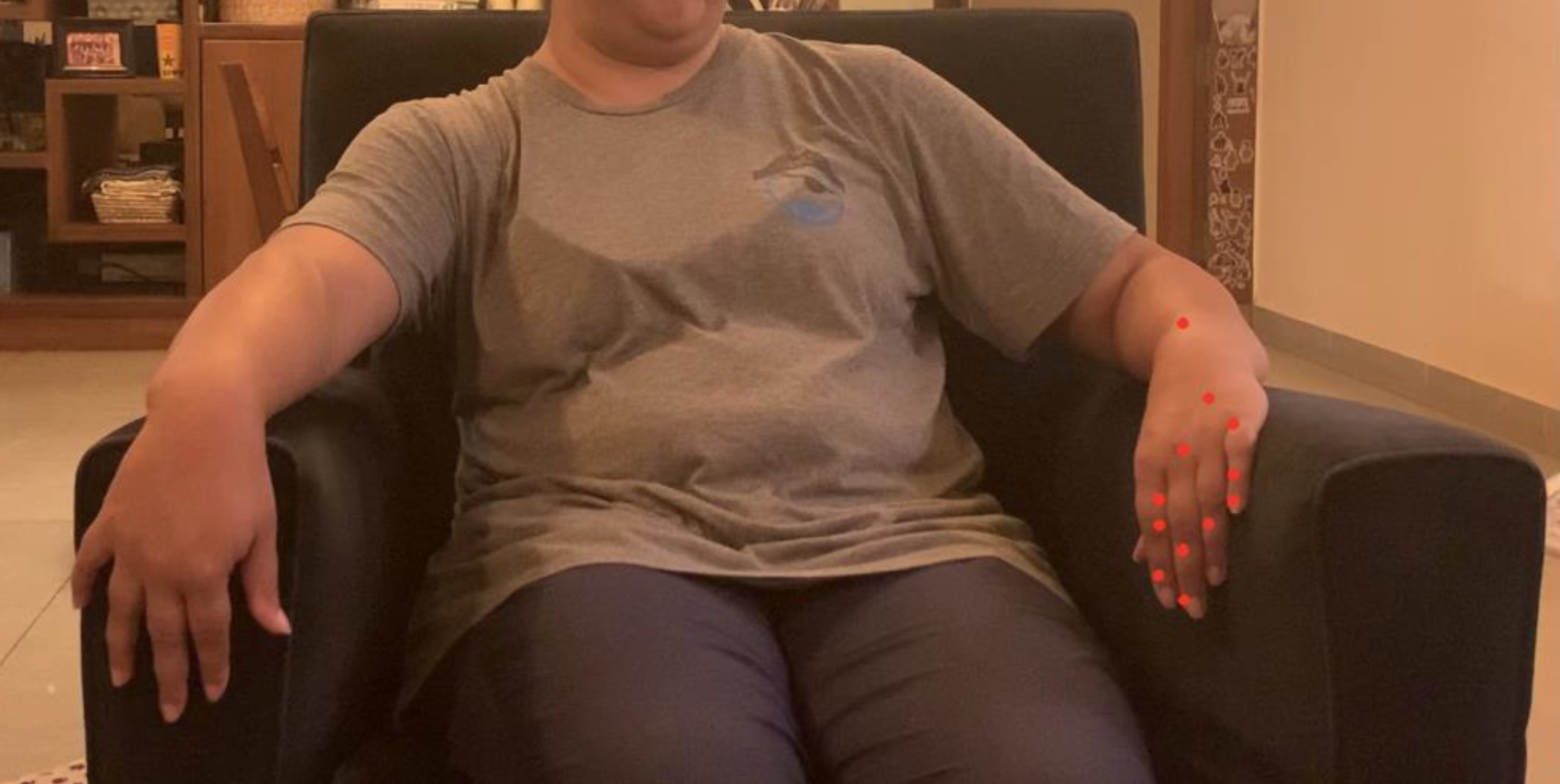}
    \caption{}
    \end{subfigure}
    \caption{Skeletonization analysis: (a) Completely incorrect detection of key points. (b) Correct detection of all 22 points. (c) Detection of incorrect hand in a busy background.}
    \label{Fig23}
\end{figure}

In this method, the skeletonization algorithm was modified to detect key points and draw circles but not draw the lines joining the key points to form the skeleton. The algorithm doesn’t show consistency in its performance. This can be attributed to the fact that the algorithm has been trained for hand pose estimation and not a still, downward-facing hand. Additionally, most hand detection and skeletonization algorithms are trained on images of the palm and the hand facing upwards rather than the downward back of the hand used in this experiment.

This algorithm has correctly detected all key points in certain cases such as Fig~\ref{Fig23}(a) and has also detected every key point incorrectly as in Fig~\ref{Fig23}(b).

It has been observed that the algorithm detects the hands’ keypoints accurately when the hand is placed flat on a surface and fully visible rather than resting on the arm of a chair as seen in Fig~\ref{Fig23}(c) where the subject’s resting hand is not detected but the key points of the hand on the chair are detected.

\subsection{Overall Results: Classification of Motions after Pre-Processing}
\label{Sec5_4}
\begin{figure}[ht]
    \centering
    \includegraphics[width = \linewidth]{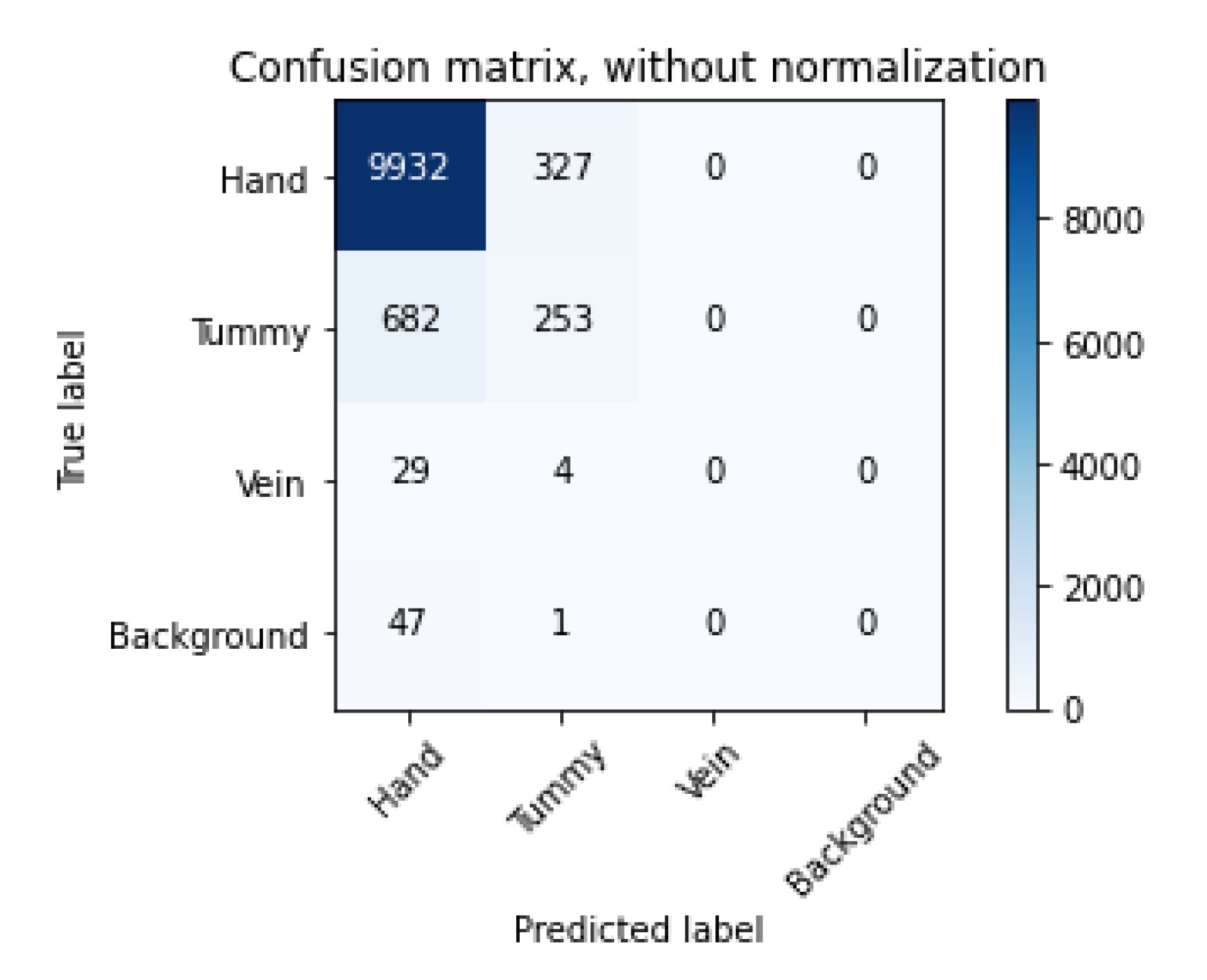}
    \caption{Confusion Matrix of kNN results.}
    \label{Fig24}
\end{figure}

The three models gained accuracies of 90.33\%, 93.37\% and 89\%. These high accuracies can be attributed to the:
\begin{itemize}
    \item large amount of data used for training,
    \item the correct selection of pixels which were artifact-free,
    \item appropriate pre-processing of raw data, and
    \item use of the best value of $k$ ($k = 3$)
\end{itemize}
However, it was also noted that the bias towards the ‘hand motion’ due to larger amounts of data, the machine tended to confuse all other categories with this motion as seen in Fig~\ref{Fig24}.

\begin{figure}[ht]
    \centering
    \includegraphics[width = \linewidth]{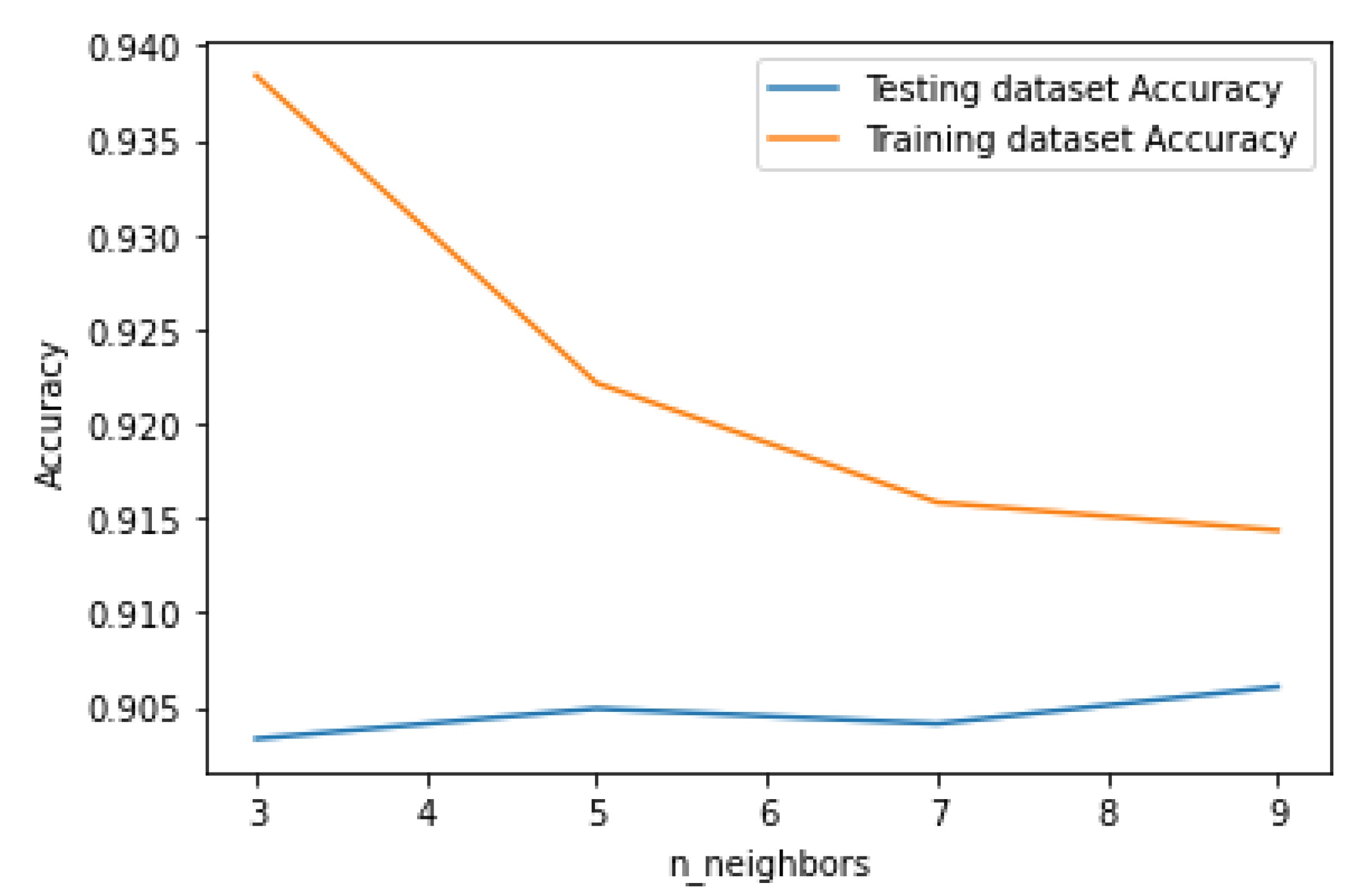}
    \caption{Accuracy of model versus $k$ nearest neighbors.}
    \label{Fig25}
\end{figure}

Overall, these results show the potential of this work to be used as a pre-processing technique after removing data biases for training. Additionally, it is observed that k value = 3 is most ideal for this use-case as it yields best accuracy (see Fig~\ref{Fig25}).

\section{Future Work}
\label{Sec6}
The work proposed in this paper discusses a method to classify the motions in the hand. Although this method is helpful to atlas the human hand’s micro-motions, there is some more useful information that can be accessed from this. For example, the tremors in the hands can be further analyzed. They can be used to aid in the remote detection of neuro-degenerative diseases and motor control diseases such as Parkinson’s Disease, Essential Tremor disease, etc. Additionally, the labeling of the stomach’s movements was done to explore the extraction of a person’s breathing rate.

Moreover, a serious challenge faced during the development of this method is the lack of data. To develop a higher level of accuracy using Deep Learning for the proposed method, a greater amount of data would help the machine learn many features. Thus, the accuracy of classification would improve. To acquire a greater amount of data, data augmentation techniques can be used in the future.

An additional observation made during this study is that many hand detection algorithms~\cite{Dibia} and skeletonization algorithms are trained on common hand poses and thus are not very accurate in detecting or skeletonizing the pose of the hand in the proposed method. Therefore, for future improvements, skeletonization and hand detection algorithms can be re-trained on the pose used for this paper for more accurate results.

\section{Conclusion}
\label{Sec7}
The computerized classification of micro-motions in the hand using waveforms from mobile phone videos applied Eulerian Video Magnification, Skeletonization, Manual Labeling and Classification using kNN to classify micro-motions from videos. Raw data were collected and collated due to lack of readily available datasets, for this project. This helped the classifier train on raw real-world data.
This project has shown the capability to be used as an advantageous pre-processing tool for micro - motion detection in the hand. However, for this, the biases in data that existed must be eliminated. Furthermore, after removal of bias and collection of sufficient data, deep learning models can also be tested for this use-case.

Ultimately, this method was able to reach an accuracy of 92.48\% on 45 videos of the hand. It is hoped this work can be improved upon to make it useful for the remote detection of motor control diseases.

\section*{Acknowledgment}
\label{Sec8}
The author would like to acknowledge that this research work was conducted as part of the Pioneer Research program and thank Prof Susan E Fox, Professor and Chair of Dept of Mathematics, Statistics and Computer Science at Macalaster College for her guidance.

\ifCLASSOPTIONcaptionsoff
  \newpage
\fi



%





\end{document}